%% file: main.tex
\documentclass[10pt,twocolumn,lettersec]{article}

\usepackage[pagenumbers]{cvpr} 

\input{preamble}

\definecolor{cvprblue}{rgb}{0.21,0.49,0.74}
\usepackage[pagebackref,breaklinks,colorlinks,allcolors=cvprblue]{hyperref}

\def\paperID{} 
\def\confName{CVPR}
\def\confYear{2026}

\title{\ours: All-Age Human Mesh Recovery}

\author{Laura Bravo-S\'anchez$^{1,2}$
\and
Matthieu Armando$^1$
\and
Romain Br\'egier$^1$
\and
Gr\'egory Rogez$^1$
\and
Serena Yeung-Levy$^2$ 
\and
Fabien Baradel $^1$
}

\begin{document}
\twocolumn[{
\renewcommand\twocolumn[1][]{#1}
\maketitle
\centering
\vspace{-3em}
\begin{center}
    $^{1}$NAVER LABS Europe \quad $^{2}$Stanford University 
\end{center}

\includegraphics[width=0.855\textwidth]{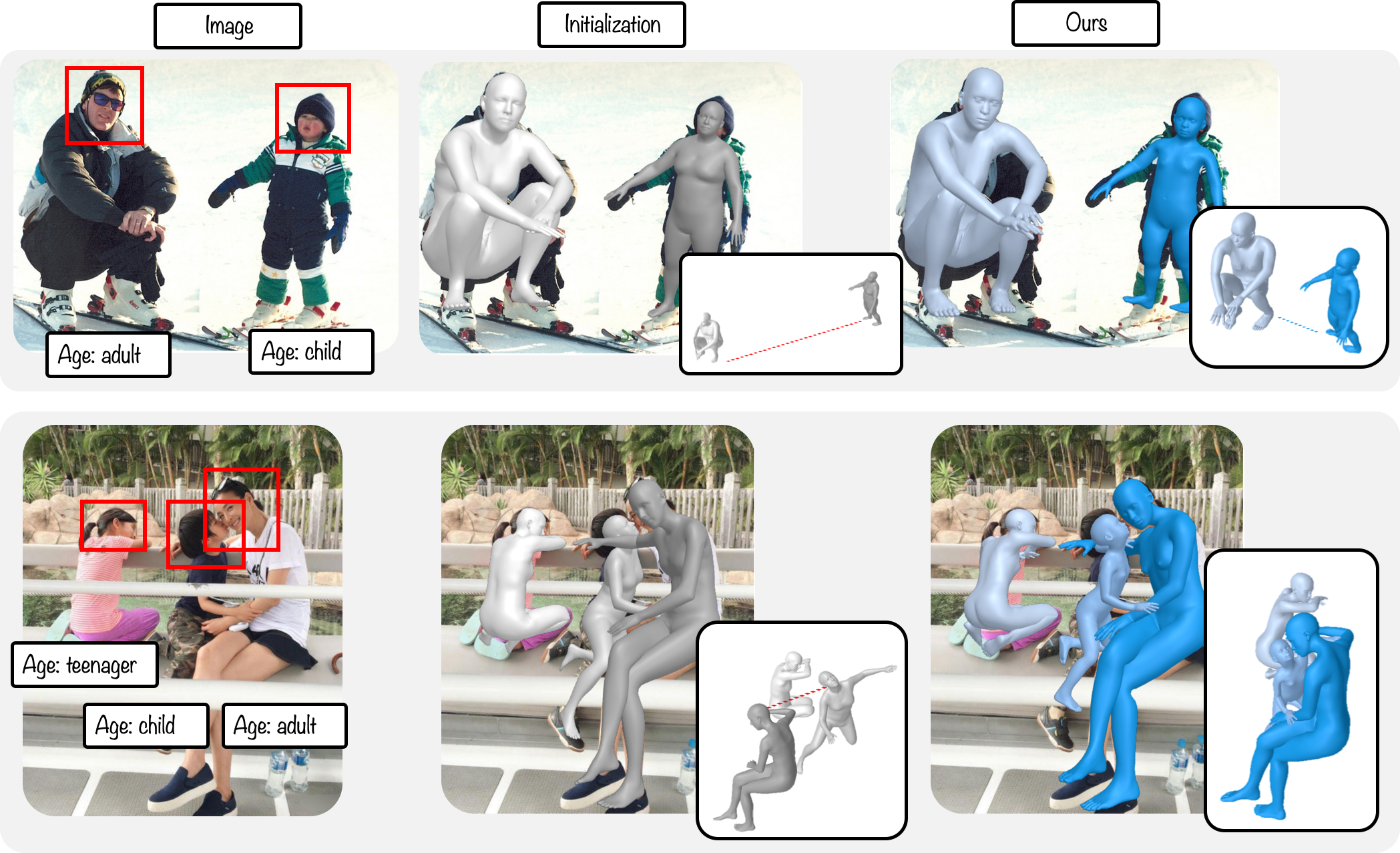}
\vspace{-2mm}
\captionof{figure}{
\textbf{\ours} recovers multi-person 3D human meshes of all ages directly in camera space. By integrating expert semantic, depth, keypoint, and segmentation cues, it improves all-age HMR and enables zero-shot adaptation of adult-only models.}
\label{fig:pull}
\vspace{2em}
}]

\maketitle
\input{sec/0_abstract}

\input{sec/1_intro}
\input{sec/2_related_work}
\input{sec/3_method}
\input{sec/4_experiments}

\input{sec/5_conclusion}

{
    \small
    \bibliographystyle{ieeenat_fullname}
    \bibliography{main}
}

\input{supplementary/supp.tex}

\end{document}

%% file: preamble.tex
\definecolor{laucolor}{RGB}{0, 102, 204}

\definecolor{fabiencolor}{RGB}{34, 139, 34}

\definecolor{romaincolor}{RGB}{102, 74, 7}

\definecolor{SeabornBlue}{HTML}{1F77B4}

\usepackage{lipsum}
\usepackage{multicol}
\usepackage{tikz,pgfplots}
\usepackage{multirow}
\usepackage{graphicx}
\usepackage{colortbl}
\usepackage{xspace}

\usepackage{xcolor}
\usepackage{tcolorbox}
\tcbuselibrary{skins}
\usepackage[accsupp]{axessibility}  

\newcommand{\ours}{Anny-Fit\xspace}

\renewcommand{\paragraph}[1]{\vspace{0.5mm}\noindent\textbf{#1}}



%% file: sec/0_abstract.tex
\begin{abstract}
\label{sec:abstract}
Recovering 3D human pose and shape from a single image remains a cornerstone of human-centric vision, yet most methods assume adult subjects and optimize each person independently. These assumptions fail in real-world, all-age scenes, where body proportions and depth must be resolved jointly. We introduce \textbf{\ours}, a multi-person, camera-space optimization framework for all-age 3D human mesh recovery (HMR). Unlike existing per-person fitting methods, \ours jointly optimizes all individuals directly in the camera coordinate system, enforcing global spatial consistency. At the core of our approach is the use of multiple forms of expert knowledge—including metric depth maps, instance segmentation, 2D keypoints, and, VLM-derived semantic attributes such as age and gender—each obtained from dedicated off-the-shelf networks. These complementary signals jointly guide the optimization, constraining the depth-scale ambiguity characteristic of all-age scenes. Across diverse datasets, \ours consistently improves 2D reprojection accuracy (+13 to 16), relative depth ordering (+6 to 7), 3D estimation error (-9 to -29) and shape estimation (+25 to +82), producing more coherent scenes. Finally, we show that VLM-based semantic knowledge can be distilled into an HMR model via the pseudo-ground-truth annotations produced by \ours on training data, enabling it to learn semantically meaningful shape parameters while improving HMR performance. Our approach bridges adult-only and all-age modeling by enabling zero-shot adaptation of adult-trained HMR pipelines to the full age spectrum without retraining. 
Code is publicly available at \url{https://github.com/naver/anny-fit}.
\end{abstract}

%% file: sec/1_intro.tex
\section{Introduction}
\label{sec:intro}

\begin{figure}[!t]
\includegraphics[width=1.0\linewidth]{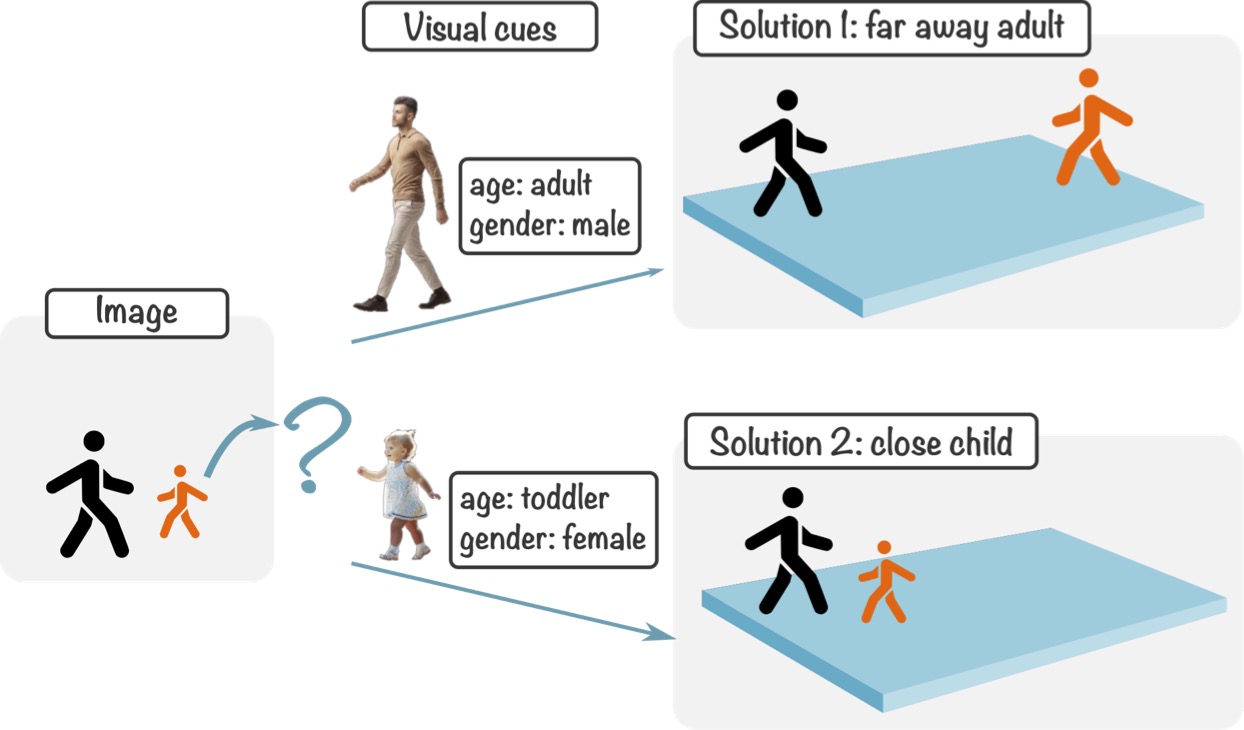}
\caption{\textbf{Depth-scale ambiguity.} Unlike adult-only settings where body size reliably indicates depth, the all-age setting generalizes the problem such that size alone cannot distinguish depth: identical 2D reprojections can correspond to either distant adults or nearby children. We leverage visual cues to infer shape and re-constrain the problem.
}
\vspace{-0.5cm}
\label{fig:shape_depth}
\end{figure}

Reconstructing 3D human pose and shape from a single monocular image (HMR) is a fundamental problem in human-centric scene understanding, pivotal for applications such as robotic navigation and embodied AI. The central challenge of this monocular task is resolving the inherent depth ambiguity: a person's apparent size in an image is a joint function of both their physical stature and their distance from the camera (Fig.\ref{fig:shape_depth}). Much prior work circumvents this ambiguity by assuming all subjects are adults \cite{zhu2020single,ugrinovic2021body}, allowing apparent size to serve as a reliable depth cue. However, this assumption breaks down in realistic, all-age settings, where the problem becomes significantly more under-constrained: a small silhouette could correspond to either a distant adult or a nearby child. Consequently, apparent size alone is insufficient, necessitating the joint estimation of  depth and body shape.

Because monocular HMR is under-specified, most approaches leverage external information to constrain the solution space. Such information may come from the scene (e.g., camera parameters~\cite{zhu2020single,baradel2024multi,zhang2025metrichmr}, person location~\cite{goel2023humans,wang2025prompthmr}, depth cues~\cite{bev}, or contact maps~\cite{muller2024generative,bravo2025ask,tripathi2023deco,muller2021self}) or from the humans themselves (e.g., 2D keypoints~\cite{kolotouros2019learning,bogo2016keep}, dense points~\cite{zhang2023pymaf,zioulis2023kbody,wang2023refit,patel2025camerahmr} or shape 
cues
~\cite{choutas2022accurate,bregier2025condimen}). Among these, we argue that the body shape cues are particularly critical for addressing the all-age ambiguity. Figure \ref{fig:shape_depth} illustrates how estimating whether a subject corresponds to an adult or a child re-constrains the depth-shape ambiguity, facilitating reliable 3D recovery.

While shape cues and an expressive body model help resolve ambiguity at the individual level, they do not by themselves guarantee \emph{scene-level} coherence in multi-person settings. Fitting each person independently to their 2D evidence often leads to inconsistent relative depths that break the scene's spatial layout. We posit that recovering a valid 3D layout requires enforcing relational depth consistency across all subjects.

To address these challenges, we introduce \textbf{\ours}, a flexible optimization framework that integrates both scene- and person-level constraints to refine initial estimates. Importantly, these constraints extend beyond human annotation: \ours leverages specialized models (such as detectors, depth estimators, and 2D keypoint regressors) together with generalist Vision-Language Models (VLMs) whose high-level semantic predictions (e.g., estimated age and gender) can be translated into meaningful shape parameters, which HMR models struggle to infer reliably. This combination enables \ours to operate either fully automatically or in a semi-supervised fashion.

A significant challenge to adapting HMR for all-age is the availability of body models that reflect different ages. While prior work~\cite{bev,patel2021agora} introduced SMPL-A to interpolate between infant and adult extremes, we adopt the more recent Anny~\cite{bregier2025human} model, which offers two key advantages for our purposes. First, Anny provides, with a single model, a continuous representation of shape variation across the full human lifespan from infants to seniors. This broader representational coverage enables consistent reasoning about shape and depth in complex multi-person all-age scenes. Second, Anny's shape space is parameterized by semantic attributes (such as age, gender, height, and weight) that align naturally with observable image cues. We exploit these two advantages with our key insight: mapping these continuous semantic attributes to discrete categories already understood by general-purpose 2D vision models. Based on this, we propose to repurpose VLMs as a training-free approach to shape estimation.

We build \ours on an optimization-based paradigm, as such methods have proven vital in HMR. Optimization methods serve dually as a post-processing step to align regression-based predictions with image evidence~\cite{zhang2023pymaf,wang2023refit} and as a generator of pseudo-ground-truth~\cite{patel2025camerahmr,muller2024generative,xia2025reconstructing} to scale training beyond small curated in-the-wild datasets and lab-controlled captures. Building on this foundation, \ours combines semantic shape initialization, joint camera-space optimization, and multi-source expert cues, enabling fully automatic reconstruction and the large-scale creation of high-quality pseudo-ground-truth for all-age, multi-person scenes.

In summary, our main contributions are:
\begin{itemize}
\item A new formulation of all-age human mesh recovery. We highlight the limitations of adult-only HMR and characterize the depth–shape ambiguity that arises in real-world all-age scenes.
\item We introduce \ours, a camera-space HMR optimization framework that jointly reconstructs multiple individuals while enforcing relational depth consistency.
\item We propose a principled fusion of expert cues
to guide all-age initialization and optimization. Our results show that \ours can improve existing all-age models and enable zero-shot adaptation of adult-only HMR models to the full age spectrum.
\item We demonstrate that \ours can generate high-quality semantic pseudo-ground truth at scale, facilitating downstream HMR models to learn semantically meaningful shape parameters while increasing accuracy.
\end{itemize}

%% file: sec/2_related_work.tex
\section{Related work}
\label{sec:related_work}

\begin{figure*}[!th]
\includegraphics[width=1.0\linewidth]{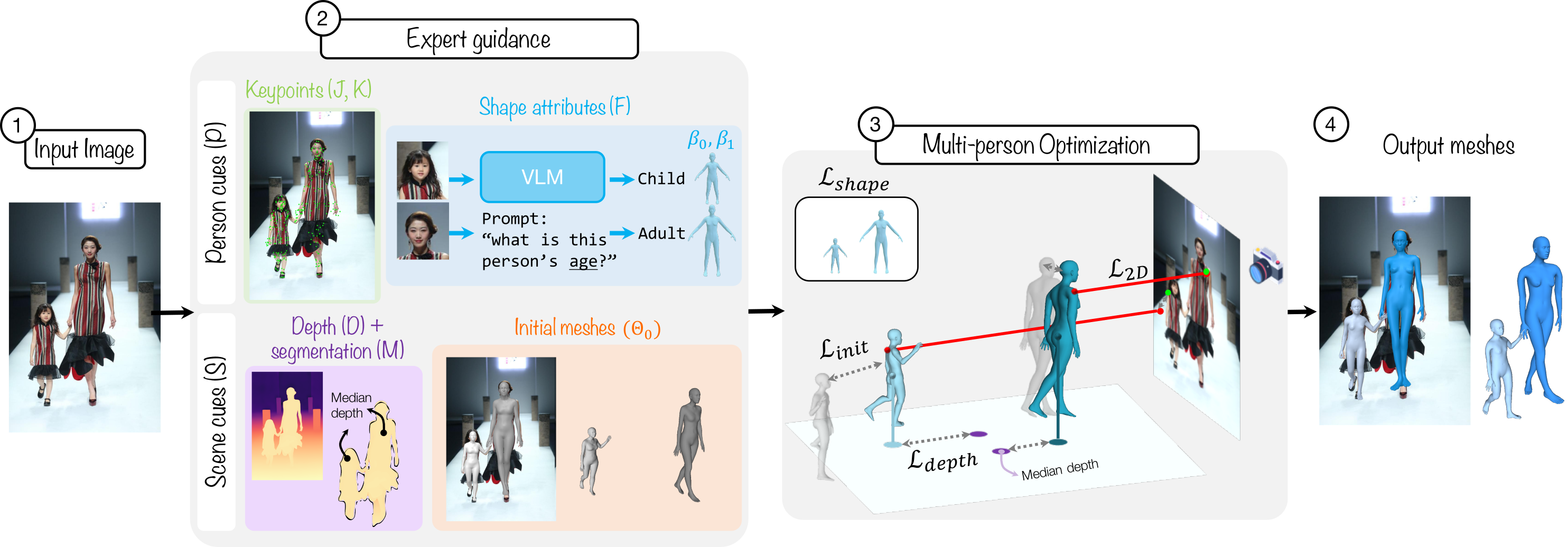}
\vspace{-1.8em}
\caption{\textbf{Overview of \ours}.
We refine initial human meshes estimated using an HMR network through iterative optimization. \ours leverages pre-computed cues from expert vision models to guide fitting. To mitigate degenerate depth solutions that satisfy 2D reprojection losses, we incorporate 
shape attribute estimation
and an explicit multi-person depth loss.}
\label{fig:method}
\vspace{-1em}
\end{figure*}

\paragraph{Multi-person HMR.}
Human Mesh Recovery (HMR) aims to estimate full 3D human bodies from monocular images~\cite{kanazawa2018end}.
Parametric methods~\cite{loper2023smpl, pavlakos2019expressive, xu2020ghum, bregier2025human} dominate this task by regressing body model parameters.
While early works focused on single-person settings~\cite{kolotouros2019learning,kanazawa2018end}, recent advances address multi-person HMR~\cite{baradel2024multi,aios,sathmr,bregier2025condimen}, where all people must be localized and recovered coherently in 3D.

Two broad paradigms exists: regression and optimization.
Regression-based methods~\cite{ROMP,bev,baradel2024multi,wang2025prompthmr,aios,sathmr} directly predict human meshes from image features, offering efficiency but limited generalization to unseen demographics or complex occlusions.
Optimization-based approaches~\cite{kolotouros2019learning, patel2025camerahmr, wang2025dtohumans} instead refine predicted mesh parameters to better match image cues, providing higher fidelity but depending heavily on initialization.
Both rely on strong priors, including pose regularization~\cite{pavlakos2019expressive, davydov2022adversarial, lu2025dposerx}, camera intrinsics~\cite{baradel2024multi}, human height~\cite{wang2025dtohumans}, or additional cues such as segmentation and text~\cite{wang2025prompthmr}.
An additional challenge in multi-person scenes is reasoning about spatial relationships.
Prior works improve relative depth via camera-space formulations~\cite{li2022cliff, baradel2024multi, patel2025camerahmr,zhang2025metrichmr}, depth supervision~\cite{bev, wang2025dtohumans}, or explicit interaction modeling~\cite{fieraru2020three, mueller2023buddi,huang2024closely,khirodkar2024harmony4d}.
In contrast, we propose a simple depth-ordering loss leveraging recent advances in monocular depth estimation to refine global 3D placement and ensure consistent camera-space positioning across all subjects given per-person shape priors.

\paragraph{Modeling age in HMR.}
Most existing HMR approaches have been designed and evaluated primarily for adults. Beyond modeling limitations, this strong adult bias also reflects data scarcity: large-scale annotated datasets containing children are extremely limited due to (i) the rarity of publicly available child imagery on the web, (ii) strict privacy and consent constraints for minors, and (iii) the ethical sensitivity surrounding the collection and release of children's images. As a result, current training corpora overwhelmingly depict adults, making it difficult for standard HMR pipelines to generalize across the full age spectrum.

Early work by Hesse \etal~\cite{hesse2018learning, hesse2018computer} introduced SMIL, a parametric model of infants (2–4 months old) derived from 3D scans.
To extend age coverage, AGORA~\cite{patel2021agora} employed SMPL-XA, a piecewise interpolation between SMPL-X~\cite{loper2023smpl,pavlakos2019expressive} and SMIL to generate synthetic child bodies, but the resulting shapes were often inconsistent due to the discontinuity between adult and infant morphologies. BEV~\cite{bev} was the first to tackle all-age estimation from in-the-wild data using weak supervision from age categories, depth layers, and 2D keypoints. HARMONI~\cite{weng2025harmoni} further explored joint modeling of SMPL and SMIL for analyzing longitudinal child growth.

Recently, Anny~\cite{bregier2025human} introduced a parametric model covering the full human lifespan, making it particularly suitable for modeling age in HMR.
However, as a new body model, Anny currently lacks the large-scale training data and broad ecosystem support available for more established models like SMPL~\cite{loper2023smpl} or SMPL-X~\cite{pavlakos2019expressive}.
This limits the effectiveness of purely regression-based learning approaches built on top of it.
To address this, we propose an optimization-based method that explicitly incorporates VLM-derived age cues to guide the fitting process, enabling reliable all-age 3D recovery in a low-data regime.

\paragraph{Language-guided human shape modeling.}
A complementary direction in HMR focuses on modeling body shape using semantic information rather than purely geometric cues. Early works model adult shape variation based on size and height~\cite{ugrinovic2021body}, while methods such as STRAPS~\cite{STRAPS2018BMVC}, BodyTalk~\cite{streuber2016body}, and SHAPY~\cite{choutas2022accurate} enrich shape estimation by leveraging textual human attributes (e.g., height, weight, body type).
SHAPY is particularly relevant to our setting as it demonstrates that semantic descriptions can guide fine-grained shape prediction from single images.  

Datasets such as SSP-3D~\cite{STRAPS2018BMVC} provide valuable 3D supervision for shape-aware prediction, yet remain limited in scale and in body-shape diversity—especially for subjects outside the adult age range—highlighting the need for alternative sources of semantic conditioning. On the other hand, facial modeling research has explored estimating human attributes such as age and other semantics directly from single images~\cite{rothe2015dex, shen2018deep}.
Parallel progress in text–image alignment has shown that language supervision provides a rich representation of human pose and shape attributes~\cite{rothe2015dex,shen2018deep} to joint embedding models that reason over visual and textual descriptions~\cite{delmas2024poseembroider,delmas_posescript,delmas2023posefix,feng2024chatpose}.
Inspired by this, we depart from training a dedicated body-shape regressor and instead leverage foundation Vision–Language Models (VLMs)~\cite{bai2025qwen2,liu2023improvedllava,liu2023llava} as general-purpose estimators of semantic human characteristics.  
By interfacing VLM predictions with the semantic shape space of the Anny body model~\cite{bregier2025human}, we translate text-derived cues directly into the shape space during optimization.
This enables flexible, data-free semantic conditioning across diverse subjects without requiring large-scale shape-labeled datasets, and provides a natural bridge between text-driven supervision and 3D body shape estimation.

%% file: sec/3_method.tex
\section{Method}
\label{sec:method}

\subsection{Problem formulation}
Our goal is to recover 3D posed human meshes of individuals of all ages from monocular RGB images. We represent each individual in the scene with the 3D parametric all-age human body model Anny~\cite{bregier2025human}, chosen for its explicit parameterization of age-dependent shape variation. The model parameters for person $i$ are $\Theta^{i} = \{\beta^{i},\phi^{i},\tau^{i},\theta^{i}\}$. 
Where $\beta^{i} \in \mathbb{R}^{10}$ denotes the shape parameters, $\phi^{i} \in \mathbb{R}^{3}$ the root orientation, $\tau^{i} \in \mathbb{R}^{3}$ the root translation, and $\theta^{i} \in \mathbb{R}^{163}$ the articulated pose. Given an input image $I$ containing $N$ individuals of varying ages, our objective is to estimate a set of parameters $\Theta_{\text{init}} = \{\Theta^{1}, \dots, \Theta^{N}\}$ from image-derived 
expert knowledge, and refine them through optimization to obtain image-aligned parameters $\Theta_{\text{final}}$.

\begin{figure*}[!t]
\centering
\includegraphics[width=0.87\linewidth]{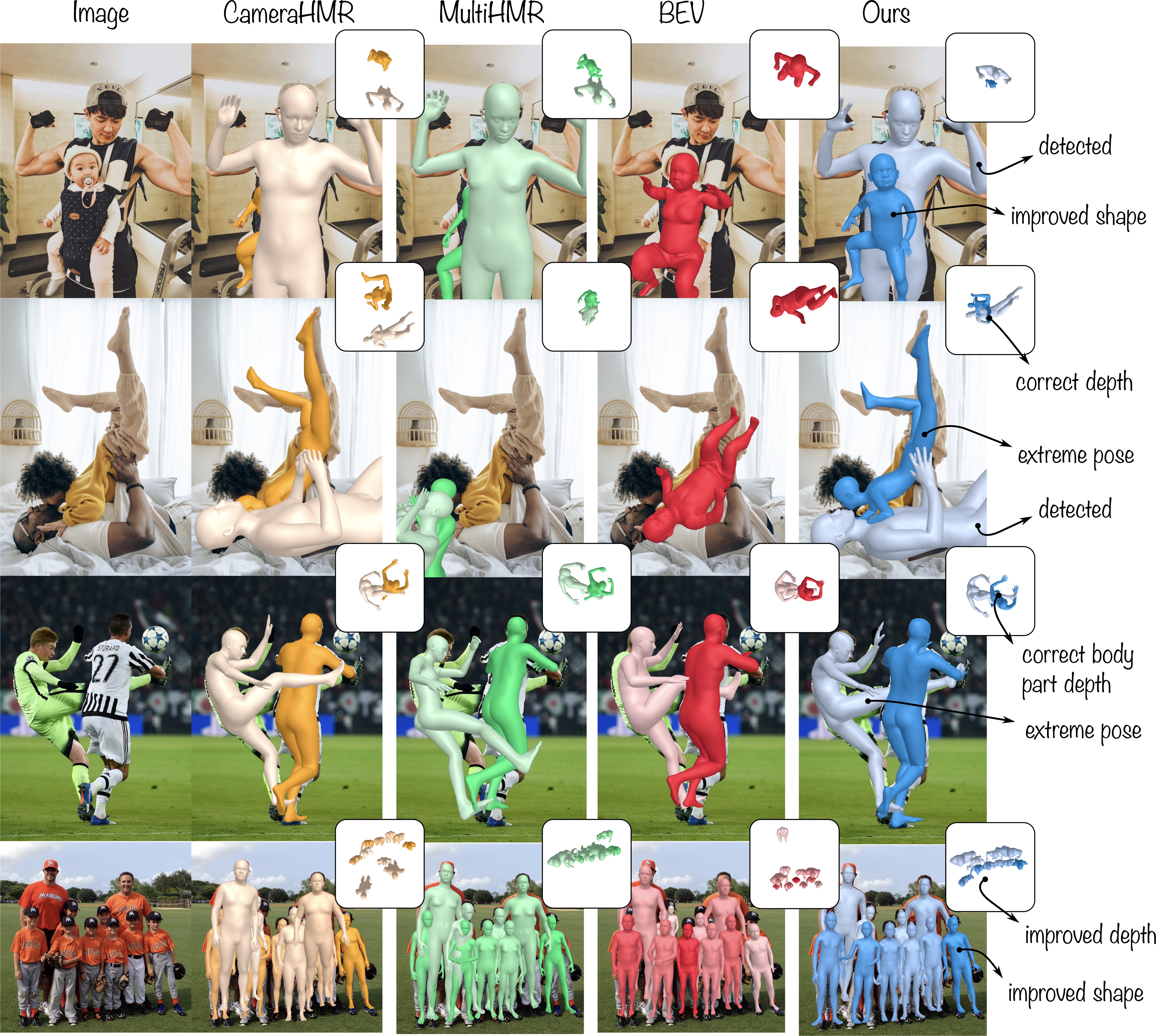}

\caption{\textbf{Qualitative results}, in front and top view. Our method, \ours, exploits the advantages of SOTA models and generalizes them to the all-age setting. Compared to BEV, this translates to improvements in detection, depth ordering, shape and pose estimation.
}
\label{fig:qual_pos}
\end{figure*}

\subsection{\ours}

Figure~\ref{fig:method} provides an overview of our approach for multi-person, all-age human mesh recovery. The core idea is to treat the task as an expert-guided optimization, where auxiliary cues from  expert models provide strong priors helping to disambiguate the mesh recovery problem.
We organize these cues into person-level $\mathcal{P}$ and scene-level 
constraints
$\mathcal{S}$, which provide weak supervision signals to guide the
fitting process. In particular, the person-level cues $\mathcal{P} = \{J, F, K\}$, consist of 2D joints location $J$, shape attributes estimates $F$ and dense 2D keypoints locations $K$. While the scene-level cues $\mathcal{S} = \{D, M, \Theta_{t-1}\}$, stem from metric depth maps ($D$) and instance segmentation ($M$) predictions, as well as the previous optimization state $\Theta_{t-1}$, used as a regularizer to prevent excessive drift during the optimization. These cues are integrated into the overall objective function, detailed below.

\paragraph{Person cues.}
Central to age estimation is the shape-depth ambiguity problem shown in Figure~\ref{fig:shape_depth}. Without a close initial shape estimate or a perfect body pose and positioning, the optimization fixes the easiest one of these often biasing the solution to an unwanted local minima.
This can happen by having a model that lacks understanding of the visual representation of a person's shape or has trouble placing a person in a scene. We address this challenge by building on the interpretable semantic shape space of the Anny model, for which each dimension of $\beta$ corresponds to non-independent physical attribute (age, gender, weight, height, muscle, etc.) and utilizing external knowledge derived from human annotation or from a foundation model.

We show how a generalist model like a VLM that has visual appearance information encoded can be queried to initialize the shape of each person. In particular, we use \cite{bai2025qwen2} as our VLM to constrain the shape optimization two-fold. First, we obtain an image-aligned estimate of $\beta$, denoted $F$, which is a mapping to the normalized Anny space of a predicted category label. Second, we use $F$ to constrain deviations throughout the optimization via a loss term $\mathcal{L}_{shape} = MSE(\beta, F)$. We query for $F$ by cropping each person’s head using $J$ when available, and otherwise by taking the detection bounding box (see Supp. Mat. for details on prompts).

We note that $F$ need not be exact but close proxy to simplify the optimization. We avoid directly regressing continuous attributes (e.g., chronological age) from the VLM, as this is notoriously difficult. Prior work~\cite{feng2024chatpose,bian2025chatgarment}, has shown the limited effectiveness of direct regression largely because it requires vast training data to overcome the limitations of tokenizers. Furthermore, such continuous attributes are often conceptually ambiguous. For instance, chronological age is an unreliable proxy for visual appearance, as two individuals of the same age can have vastly different biological ages and, consequently, different body shapes. Given these challenges, our key insight is to reformulate shape cue estimation as a categorization task over each dimension of $\beta$ (which has been shown to be more effective for VLM prediction~\cite{zhao2023survey,du2024teach}). We select anchor points with semantic meaning and remaping each prediction in the normalized parameter space of $\beta$.
In our experiments, we focus on two primary axes of variation—age and gender—while noting that this process can be naturally extended to capture a wider range of body shapes. Namely for age we select six anchor points covering the human life-span, with more detailing on early ages where shape has rapid change: 'baby', 'toddler', 'child', 'teenager', 'adult', and 'senior'. For gender we select 3 anchors: 'male', 'neutral' and, 'female'. For all attribute types we set a fallback 'unknown' as the center of the range.

In addition, we rely on $J$ and $K$, which are both sets of 2D points $p_j$ to guide optimization to the image. We assume that there is a known direct one-to-one correspondence between each $p_j$ and some 3D point $q_j$ attached to the body model.
Our losses are then $\mathcal{L}_{2D} = \mathcal{L}_{dense} = \frac{1}{|V|} \sum_{j \in V} \rho(c_j \| \hat{p}_j - p_j \|_2, \sigma)$, where $\hat{p}_j$ = $\Pi(q_j)$ is the reprojection of $q_j$ on the image plane, using known or estimated camera intrinsics.
$c_j \in [0, 1]$ is the confidence of point $j$, $V$ is the set of points ($J$ or $K$)
and $\rho(x, \sigma) = \frac{\sigma^2 x^2}{\sigma^2 + x^2}$ is the Geman-McClure robust error function to handle outlier values. We obtain $J$ and $K$ from estimators \cite{xu2022vitpose+} and \cite{patel2025camerahmr} respectively.

\input{tables/table_sota_merged}

\paragraph{Multi-person optimization.}
To obtain $\Theta_{\text{final}}$, we perform a full-scene, multi-stage optimization that refines the set of meshes $\Theta_{t}$ jointly across all individuals.
Unlike prior approaches that optimize per-person crops and remap to image coordinates, our formulation operates directly in 3D space, enforcing relational consistency between subjects and avoiding independent local minima. The optimization proceeds in stages to prevent degenerate solutions: we first optimize only translation $\tau$ to resolve coarse depth placement, then optimize $\{\tau,\phi,\beta\}$ to refine global orientation and shape attributes while preserving stable positioning, and finally optimize all parameters $\{\tau,\phi,\beta,\theta\}$ to recover detailed pose.

\textbf{Scene cues}
To place all individuals within a coherent scene, we incorporate spatial relationships by extending the depth ordering loss from~\cite{bev}, denoted $L_{depth}$, to continuous pseudo ground-truth depth values. This loss encourages people predicted to lie on the same depth plane to be close together, while separating those on different planes. While~\cite{bev} relies on manually annotated depth levels, we instead make use of an off-the-shelf metric depth map estimator~\cite{piccinelli2025unidepthv2}, which we found to provide consistent ordinal depth cues. We estimate the median depth of each person using the corresponding segmentation mask $M$ predicted using~\cite{ren2024grounded},
which we find more robust than relying on $J$ due to potential occlusions and errors in $D$. In addition to depth ordering, we further stabilize the optimization by regularizing with the previous estimates $\Theta_{t-1}$, ensuring smooth refinement from the initialization $\Theta_{init}$.

In sum, the weighted loss function for our \ours optimization is:
\begin{align*}
    \mathcal{L} ={} & \lambda_{2D}\mathcal{L}_{2D} + \lambda_{dense}\mathcal{L}_{dense} + \lambda_{shape}\mathcal{L}_{shape} + \\
     & \lambda_{init}\mathcal{L}_{init} + \lambda_{depth}\mathcal{L}_{depth}
\end{align*}
where the $\lambda$ coefficients balance the contributions of the individual losses, which are also flexibly adjusted across optimization stages (see Supp. Mat for more details).

%% file: tables/table_sota_merged.tex
\begin{table*}[th]
\centering
\caption{\textbf{Reconstruction performance.} Our method consistently improves both all-age and adult-only initializations across all metrics by a large margin, with adult-only models becoming competitive with all-age methods. $\Delta$: improvement over initialization.}
\label{tab:sota_combined}
       \begin{subtable}{0.55\linewidth}
        \centering
        \caption{\textbf{In-the-wild reconstruction on Relative Human test.} $*$: trained on the train set.}
        \label{tab:results_sota}
        \resizebox{\linewidth}{!}{%
        \begin{tabular}{r|cccccccc} 
        \toprule
        \multicolumn{1}{r|}{\multirow{2}{*}{Method}} & \textbf{2D} ($\uparrow$) & \multicolumn{5}{c}{$\mathbf{PCRD^{0.2}}$ ($\uparrow$)} & \multicolumn{1}{c}{\textbf{Age} ($\uparrow$)} & \multicolumn{1}{c}{\textbf{Gender} ($\uparrow$)} \\
        \multicolumn{1}{c|}{} & $mPCKh^{0.6}$ & overall & adult & teen & kid & baby & F1 & F1 \\ \midrule
        \textit{BEV*}~\cite{bev} & \textit{74.78} & \textit{69.19} & \textit{70.65} & \textit{68.27} & \textit{65.35} & \textit{61.96} & \textit{30.32} & \textit{0.00}\\
        \midrule
        \rowcolor[HTML]{F2F2F2} Multi-HMR~\cite{bregier2025human} & 65.39 & 59.79 & 60.73 & 64.58 & 56.45 & 46.3 & 23.29 & 34.83 \\
        
        + \textbf{Ours} & 78.84 & 66.11 & 66.92 & 65.59 & \textbf{67.08} & \textbf{57.21} & 48.57 & 81.11 \\
        
        $\Delta$ & {\color[HTML]{00B050} +13.45}  & {\color[HTML]{00B050} +6.32} & {\color[HTML]{00B050} +6.19} & {\color[HTML]{00B050} +1.01} & {\color[HTML]{00B050} +10.63} & {\color[HTML]{00B050} +10.91} & {\color[HTML]{00B050} +25.28} & {\color[HTML]{00B050} +46.28} \\
        
        \rowcolor[HTML]{F2F2F2} CameraHMR~\cite{patel2025camerahmr} & 64.69 & 59.59 & 59.80 & 67.49 & 46.59 & 30.71 & 0.00 & 0.00 \\
        + \textbf{Ours} & \textbf{81.06} & \textbf{67.24} & \textbf{67.74} & \textbf{70.36} & 65.60 & 51.09 & \textbf{48.75} & \textbf{82.13} \\
        
        $\Delta$ & {\color[HTML]{00B050} +16.37} & {\color[HTML]{00B050} +7.65} & {\color[HTML]{00B050} +7.94} & {\color[HTML]{00B050} +2.87} & {\color[HTML]{00B050} +19.01} & {\color[HTML]{00B050} +20.38} & {\color[HTML]{00B050} +48.75} & {\color[HTML]{00B050} +82.13} \\
        \bottomrule
        \end{tabular}
        } %
    \end{subtable}
    \hfill
    \begin{subtable}{0.4074\linewidth}
        \centering
        \caption{\textbf{3D reconstruction on CMU panoptic-toddler.}}
        \label{tab:cmu}
        \resizebox{\linewidth}{!}{%
        \begin{tabular}{r|cc|cc}
        \toprule
        \multirow{2}{*}{\textbf{Method}} & \multicolumn{2}{c|}{\textbf{Root}} & \multicolumn{2}{c}{\textbf{Joint-PA}} \\
         & MPJPE ($\downarrow$ mm) & PCK ($\uparrow$ \%) & MPJPE ($\downarrow$ mm) & PCK ($\uparrow$ \%) \\
         \midrule

        AiOS \cite{aios} & 162.39 & 54.15 & 723.14 & 7.78 \\
        SAT-HMR \cite{sathmr} & 153.88 & 56.40 & 654.87 & 5.50 \\
        \rowcolor[HTML]{F2F2F2} Multi-HMR \cite{bregier2025human} & 102.15 & 84.50 & 263.78 & 41.89 \\
        + \textbf{Ours} & \bf{92.52} & \bf{85.50} & \bf{223.13} & \bf{45.12} \\
        $\Delta$ & {\color[HTML]{00B050} -9.63} & {\color[HTML]{00B050} +1.00 } & {\color[HTML]{00B050} -40.66 } & {\color[HTML]{00B050} +3.23 } \\
         \rowcolor[HTML]{F2F2F2} CameraHMR \cite{patel2025camerahmr} & 149.52 & 60.18 & 658.90 & 5.53 \\
        + \textbf{Ours} & 119.93 & 74.41 & 348.03 & 23.32 \\
        $\Delta$ & {\color[HTML]{00B050} -29.60} & {\color[HTML]{00B050} +14.23 } & {\color[HTML]{00B050} -310.86} & {\color[HTML]{00B050} +17.79 } \\
        \bottomrule
        \end{tabular}%
        }
    \end{subtable}
\vspace{-.9em}
\end{table*}

%% file: sec/4_experiments.tex
\section{Experiments}
\label{sec:experiments}

\paragraph{Initialization.}
To validate our optimization-based framework, we initialize it with two complementary feedforward methods: Multi-HMR~\cite{baradel2024multi,bregier2025human} and CameraHMR~\cite{patel2025camerahmr}. 
Multi-HMR uses the Anny model and is trained only on synthetic data (Anny-One~\cite{bregier2025human}), whereas CameraHMR is a detection-based, single-person, adult-only model trained on a mixture of synthetic data and real-world pseudo-ground-truth fits.
Multi-HMR~\cite{bregier2025human} outputs Anny body parameters directly, while for CameraHMR we fit Anny parameters to the SMPL meshes~\cite{loper2023smpl}. For both initial predictions we use bounding boxes, for the bottom-up method Multi-HMR we force predictions using the closest patch prediction of the estimated nose keypoint.

\paragraph{Evaluation benchmarks and metrics.} We compare to prior work on the in-the-wild all-age dataset Relative Human~\cite{bev} and evaluate 3D recovery on the 5 toddler sequences of the CMU Panoptic dataset~\cite{Joo_2017_TPAMI} as well as in Hi4D~\cite{yin2023hi4d}.
For 2D evaluation, we use mean Percentage of Correct Keypoints ($mPCK^{0.6}$), Percentage of Correct Depth Relations ($PCDR^{0.2}$), F1 for bounding-box detection, and age and gender prediction.
For 3D evaluation we measure MPJPE on matched joints and PCK $@15cm$ to account for unmatched joints.
We consider both the per-person root aligned and Procrustes alignment (i.e. Joint-PA) for all people as one to account for inter-person accuracy~\cite{mueller2023buddi}.
We include more details in the Supp. Mat.

\paragraph{Comparison to state-of-the-art}
Table~\ref{tab:results_sota} reports results on the Relative Human test and comparison with existing methods.  \ours provides substantial improvements over zero-shot initializations across all metrics, in several cases matching or surpassing BEV, the current SOTA, despite BEV being trained directly on the dataset.
The results also highlight the coupled nature of the task: improvements in shape attribute estimation (age and gender F1) directly reduce depth–shape ambiguity, enabling more reliable multi-person depth ordering. This stronger spatial grounding, in turn, supports more accurate pose recovery as reflected by 2D reprojection gains. These findings are further supported by the 3D evaluation on CMU-Toddler reported in Table~\ref{tab:cmu}, which measures reconstruction accuracy on real-world  multi-age sequences.
\ours improves single-person pose recovery (root-aligned metrics) and multi-person relative placement (joint-PA metrics).

\input{tables/ablations_method}
\input{tables/ablations_vlm}

\begin{figure}[!t]
\includegraphics[width=1.0\linewidth]{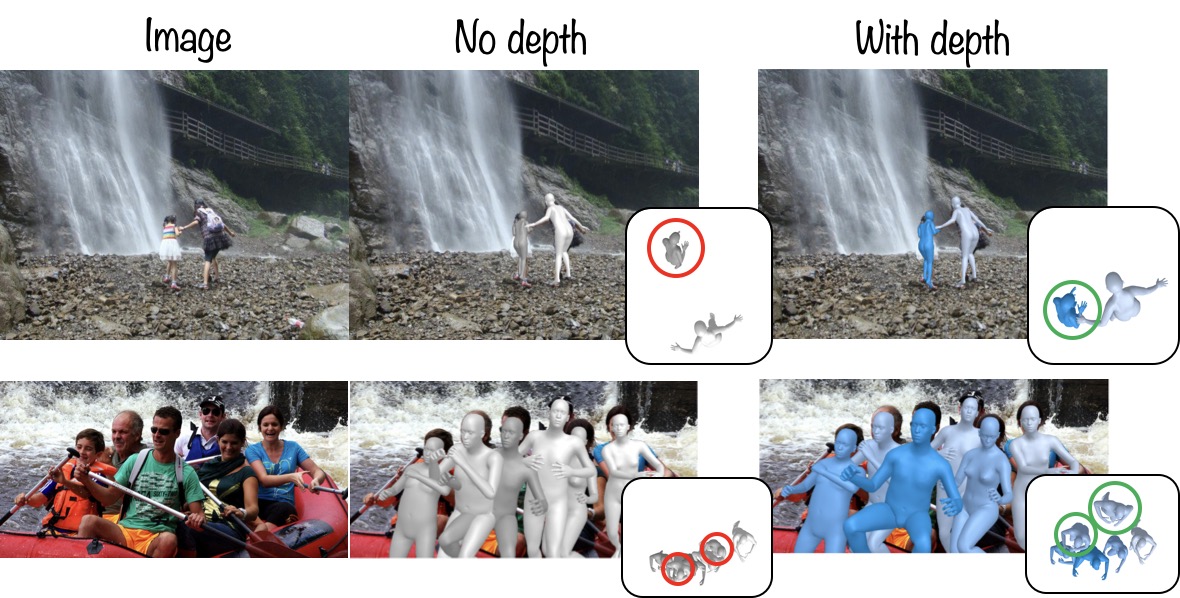}
\vspace{-2em}
\caption{\textbf{Effect of depth loss.}
Adding a depth-based loss from an expert model preserves the relative depth relationships between people. All results use the same initialization and shape prediction. Circles denote incorrect (\textcolor{red}{red}) and correct (\textcolor{ForestGreen}{green}) placement.}
\label{fig:qual_depth}
\vspace{-1em}
\end{figure}

\begin{figure}[!t]
\includegraphics[width=1.0\linewidth]{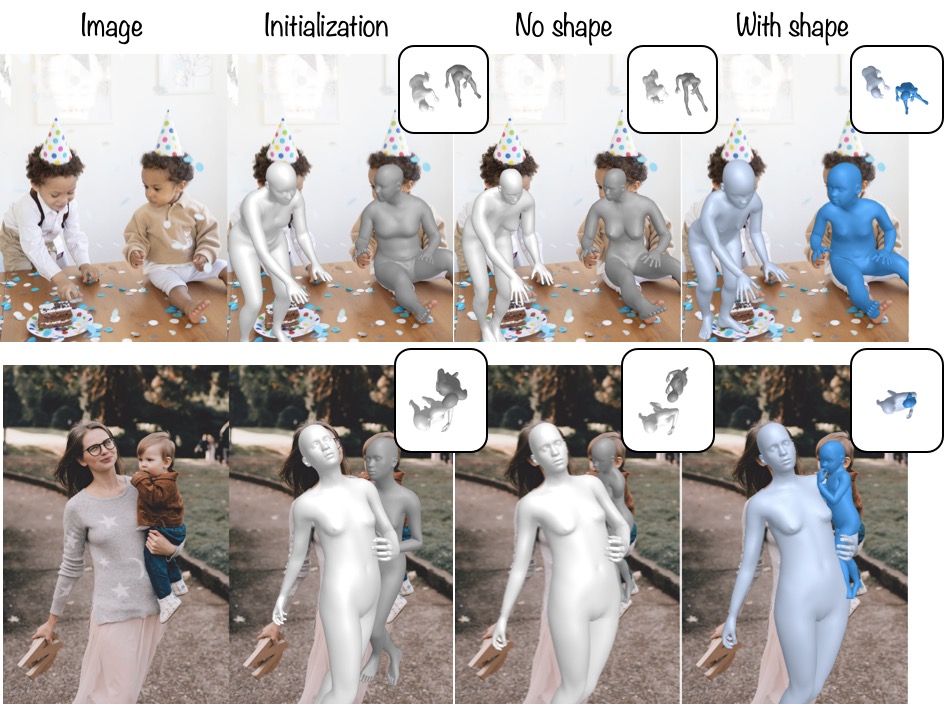}
\caption{\textbf{Effect of shape.} Top: Incorrect shape initialization
prevents the optimization from converging. Bottom: Accurate shape initialization resolves depth ordering.}
\label{fig:qual_shape}
\vspace{-0.8em}
\end{figure}

Overall, we demonstrate that \ours consistently benefits different model types. For the specialized all-age model (Multi-HMR), improvements stem from better alignment between subject appearance and 3D shape, mitigating limitations of its synthetic training distribution. At the same time, \ours shows that predictions from a non-specialized, adult-only model (CameraHMR) can be adapted to the all-age setting in a training-free manner, achieving competitive performance with specialized methods.

Figure~\ref{fig:qual_pos} further visualizes these capabilities on challenging in-the-wild scenes featuring high shape variance, extreme poses, and close interactions. \ours leverages and generalizes SOTA capabilities; it refines relative depth ordering in both multi-age arrangements (rows 1, 4) and adult-only scenes (row 3).
By building upon accurate pose estimators such as CameraHMR (rows 2, 3), \ours captures extreme poses, and using a stronger detectors can increase recall in certain scenes with challenging poses or heavy occlusions (rows 1, 2).
Finally, by using the more representative Anny body model, \ours avoids shape artifacts common in SMPL-A methods such as BEV, where children may appear as oversized infants (row 2). These results highlights the flexibility of \ours, which directly benefits from advances in HMR methods even when they are not designed for all-age reconstruction.

\subsection{Ablation experiments}
We study each component of our method on the Relative Human validation set.
To counter the over-representation of adults, we report on a 'has child' subset containing all images with at least one non-adult (see Supp. Mat.). 

\paragraph{Component ablation study.}
Table~\ref{tab:ablations_method} summarizes the contributions of individual components.
Adding weak 2D keypoint supervision improves $mPCK^{0.6}$ and yields small gains in shape, but has limited effect on depth.
In contrast, depth supervision provides a large boost to $PCDR^{0.2}$, especially for CameraHMR initializations where non-adult predictions are severely miss-scaled.
The depth loss is crucial for bringing these cases back into a valid scale (see Fig.~\ref{fig:qual_depth}). 

\input{tables/trained_fits}

Integrating VLM-estimated shape attributes --- both at initialization and as a shape loss --- has the largest impact, producing substantial gains in age and gender classification (+30 and +40 F1, respectively).
It also improves depth and pose accuracy.
As shown in Fig.~\ref{fig:qual_shape}, when the initial shape is far from the true one, optimization cannot recover it without shape guidance.
Consistent with prior work ~\cite{bev,ugrinovic2021body} on the depth-scale ambiguity, using the correct shape leads to more reliable relative depth. We further evaluate two depth losses: a learned affine transformation of the per-person predicted median depth (RD) and the final relative depth ordering loss (D). Both improve multi-person depth consistency, with the ordering loss performing best overall.
The full model, combining all components, achieves the strongest overall performance across metrics, underscoring the necessity of jointly modeling shape, depth, and pose in the all-age setting.

\begin{figure}[!ht]
    \centering
    \includegraphics[width=1.0\linewidth]{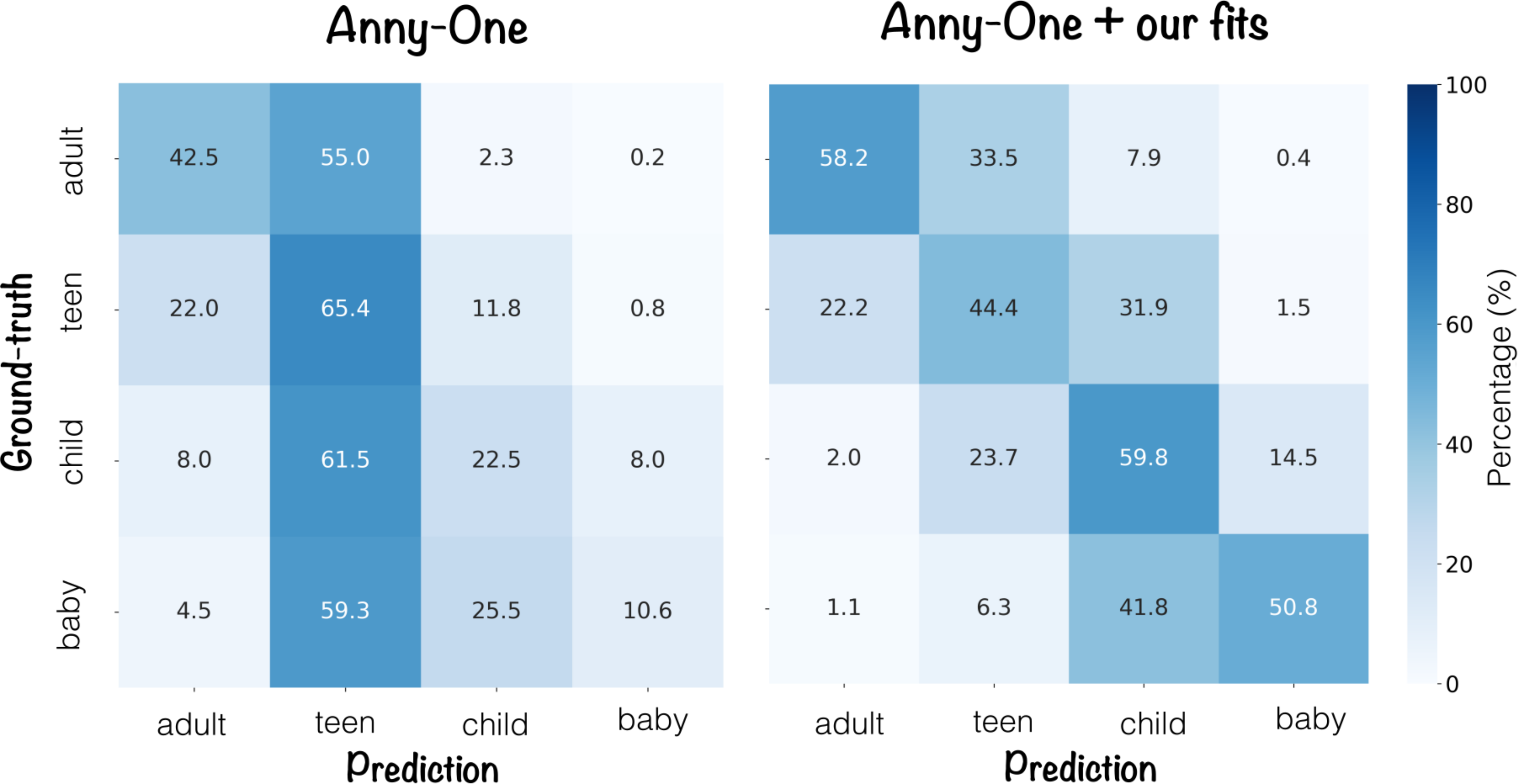}
    \caption{\textbf{Age confusion matrix on Relative Human test.} After retraining with our fits, age alignment improves.}
    \label{fig:conf_mat}
\end{figure}

\paragraph{VLM shape attribute estimation.}
Table~\ref{tab:vlm_res} reports zero-shot shape-attribute predictions from open-source VLMs that fit on a single H100 GPU, using prompt aligned with Relative Human's class definitions.
Gender is predicted accurately across all models, suggesting that this attribute is well captured.
In contrast, age estimation shows large variability: performance is strongest for adults, while most errors occur between neighboring age ranges, indicating that finer age distinctions remain challenging.

\paragraph{Pseudo-ground truth generation.}
Finally, we assess whether our high-quality fits can be used to train feedforward HMR methods.
To this end, we processed 30k images from the MS-COCO~\cite{lin2014microsoft} training set, which provides diverse, in-the-wild scenes spanning a broad range of ages.
We trained Multi-HMR using various combinations of synthetic data, our optimized fits, and fits generated by the optimization method CamSimplify~\cite{patel2025camerahmr}.
All models are trained for 600K steps at input resolution of $672\times672$.

We report results in Table~\ref{tab:retrain}.
On Relative Human, adding our fits mitigates the limitations of synthetic-only training and yields large improvements in shape estimation (+18, +47), 2D reprojection (+7), and depth ordering (+5–13). Fig.~\ref{fig:conf_mat} illustrates how our fits improve alignment of predicted age after training.
In contrast, adding CamSimplify fits degrades performance, showing that gains stem from the quality of the pseudo-GT rather than simply more data.
Results on Hi4D confirm the same trend: our fits provide larger improvements than CamSimplify even on an adult-only dataset.
Overall, these findings highlight the importance of high-quality, camera-consistent multi-person fits and demonstrate the potential of \ours to enhance a base model when full supervision is limited.

\begin{figure}[!t]
\includegraphics[width=1.0\linewidth]{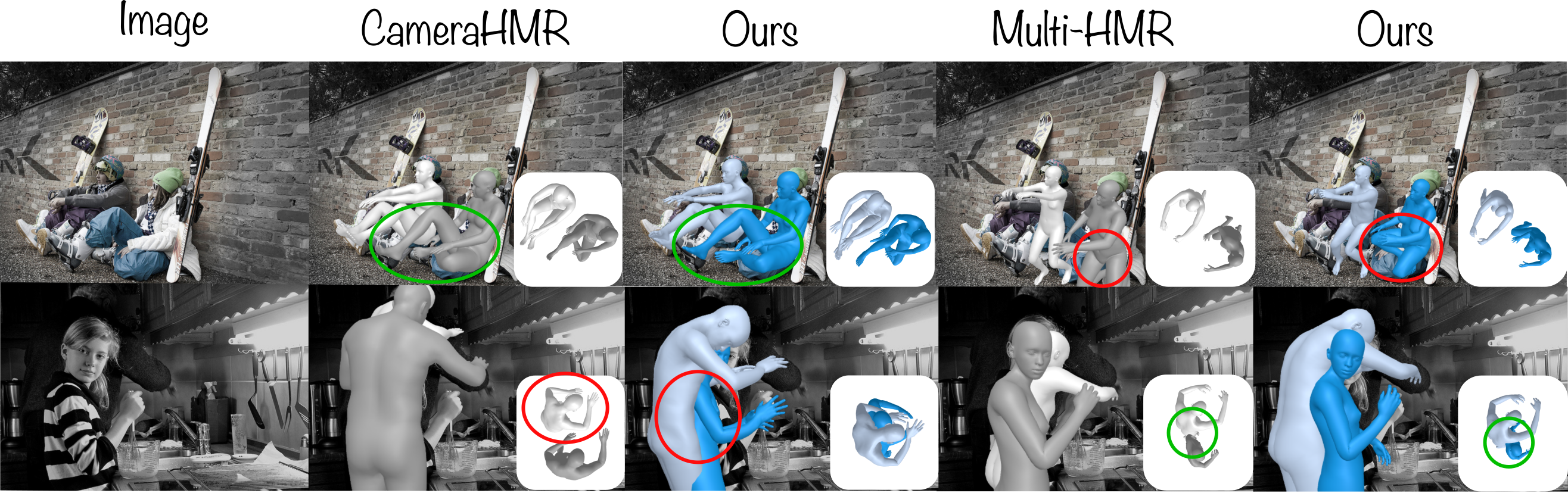}
\vspace{-1.8em}
\caption{\textbf{Example of failure cases}. Top: pose of legs, legs usually have low keypoint confidence. Bottom: interpenetration because of initial position and not enough guiding keypoints.
}
\label{fig:qual_neg}
\vspace{-1em}
\end{figure}

%% file: tables/ablations_method.tex
\begin{table}[!ht]
\centering
\caption{\textbf{Method ablation study.} \ours main component analysis on the Relative Human validation 'has child' subset.
\colorbox[HTML]{FFF2CC}{\phantom{x}} indicates the default setting.
O: Multi-person optimization, S: VLM shape, D: depth, RD: root depth.
}
\vspace{-.7em}
\label{tab:ablations_method}
\resizebox{\linewidth}{!}{%
\begin{tabular}{l|c|ccccc|cc}
\toprule
    \multicolumn{1}{c|}{\multirow{2}{*}{\textbf{Method}}} & \textbf{2D} & \multicolumn{5}{c|}{{\textbf{$\mathbf{PCRD^{0.2}}$}}} & \multicolumn{1}{c}{\textbf{Age}} & \multicolumn{1}{c}{\textbf{Gender}} \\
    \multicolumn{1}{c|}{} & $mPCKh^{0.6}$ & overall & adult & teen & kid & baby & F1 & F1 \\
    \midrule
    \rowcolor[HTML]{F2F2F2} 
    Multi-HMR \cite{bregier2025human} & 60.99  & 62.66 & 61.22 & 77.10 & 62.85 & 56.07 & 28.55 & 41.95 \\
    + O & 71.76  & 60.09 & 59.47 & 69.73 & 59.33 & 56.82 & 36.05 & 77.41 \\
    + O + S & 76.28 & 59.95 & 61.37 & 68.83 & 58.16 & 58.28 & 59.70 & 84.53 \\
    + O + D & 79.21 & 63.37 & 63.21 & 72.45 & 63.94 & 54.17 & 30.61 & 38.38 \\
    + O + S + RD & 78.22 & 64.86 & 66.55 & 71.52 & \bf{63.59} & \bf{60.79} & \bf{57.45} & \bf{84.91} \\
    \rowcolor[HTML]{FFF2CC} 
    + O + S + D & \bf{79.22} & \bf{65.13} & \bf{66.67} & \bf{72.99} & 63.31 & 58.86 & 56.78 & 83.75 \\
    \midrule
    \rowcolor[HTML]{F2F2F2} 
    CameraHMR \cite{patel2025camerahmr} & 43.84 & 50.82 & 46.41 & 66.12 & 52.87 & 34.06 & 18.49 & 38.04 \\
    + O & 79.09 & 55.01 & 50.92 & 71.33 & 55.88 & 38.22 & 22.47 & 39.16 \\
    + O +  S & 80.90 & 55.78 & 53.69 & 64.85 & 57.82 & 42.45 & \bf{57.45} & \bf{83.45} \\
    + O + D & 79.86 & 59.43 & 57.30 & 68.68 & 61.38 & 45.91 & 43.94 & 41.27 \\
    + O + S + RD & \bf{82.27} & 66.15 & 62.43 & 77.23 & 68.45 & \bf{58.47} & 56.99 & 82.09 \\
    \rowcolor[HTML]{FFF2CC} 
    + O + S + D & 81.62 & \bf{67.55} & \bf{66.06} & \bf{78.87} & \bf{70.65} & 53.77 & 56.52 & 81.43 \\
    \bottomrule
\end{tabular}
}
\vspace{-.5em}
\end{table}

%% file: tables/ablations_vlm.tex
\begin{table}[!th]
\centering
\caption{\textbf{Shape attribute performance.} Results of varying VLM models on the Relative Human validation 'has child' subset.
\colorbox[HTML]{FFF2CC}{\phantom{x}} indicates the default setting.
}
\vspace{-.9em}
\label{tab:vlm_res}
\resizebox{\linewidth}{!}{%
\begin{tabular}{l|ccccc|c} 
\toprule
\multicolumn{1}{c |}{\multirow{2}{*}{\textbf{Method}}} & \multicolumn{5}{c |}{\textbf{Age} \small F1} & \textbf{Gender} \\
\multicolumn{1}{c |}{} & overall & adult & teen & kid & baby & F1 \\
\midrule
SmolVLM-Instruct ~\cite{marafioti2025smolvlm} & 43.06 & 28.05 & 34.46 & 52.31 & 57.41 & 90.23 \\
ViP Llava 13B~\cite{cai2024vipllava} & 46.64 & 62.44 & \bf{62.98} & 19.23 & 41.89 & 83.87 \\
Qwen2.5 VL 3B~\cite{bai2025qwen2} & 63.57 & 81.58 & 36.49 & 66.34 & \bf{69.86} & \bf{92.26} \\
\rowcolor[HTML]{FFF2CC}
Qwen2.5 VL 7B~\cite{bai2025qwen2} & \bf{67.23} & \bf{85.98} & 51.57 & \bf{70.29} & 61.08 & 92.03 \\
\bottomrule
\end{tabular}
}
\vspace{-0.3cm}
\end{table}

%% file: tables/trained_fits.tex
\begin{table*}[!t]
    \centering
    \caption{\textbf{Training feedforward models with pseudo ground-truth}. Comparing Multi-HMR retrained with different data.}
    \label{tab:retrain}
    \begin{subtable}{0.6\linewidth}
            \centering
            \caption{\textbf{Relative Human test.}}
    \label{tab:retrain_rh}
    \resizebox{\linewidth}{!}{%
    \begin{tabular}{l|c|ccccc|cc}
\toprule
\multicolumn{1}{c|}{\multirow{2}{*}{\textbf{Data}}} & \textbf{2D} & \multicolumn{5}{c|}{\textbf{$\mathbf{PCRD^{0.2}}$}} & \multicolumn{1}{c}{\textbf{Age}} & \multicolumn{1}{c}{\textbf{Gender}} \\
 & \small $mPCKh^{0.6}$ & overall & adult & teen & kid & baby & F1 & F1 \\ \midrule
\rowcolor[HTML]{F2F2F2}  Anny-One~\cite{bregier2025condimen} & 62.70 & 63.42 & 63.93 & 71.40 & 59.70 & 44.17 & 24.47 & 33.60 \\
Anny-One + \cite{patel2025camerahmr} fits & 59.75 & 52.09 & 53.10 & 59.58 & 43.17 & 25.93 & 11.75 & 29.55 \\
Anny-One + \textbf{our fits} & \textbf{70.18} & \textbf{68.68} & \textbf{69.03} & \textbf{76.83} & \textbf{66.42} & \textbf{57.52} & \textbf{42.96} & \textbf{81.32} \\
$\Delta$ & {\color[HTML]{00B050} +7.48} & {\color[HTML]{00B050} +5.26} & {\color[HTML]{00B050} +5.10} & {\color[HTML]{00B050} +5.43} & {\color[HTML]{00B050} +6.72} & {\color[HTML]{00B050} +13.35} & {\color[HTML]{00B050} +18.49} & {\color[HTML]{00B050} +47.72} \\
\bottomrule
\end{tabular}
    }
    \end{subtable}
    \hspace{1em}
    \vspace{1em}
    \centering
    \begin{subtable}{0.33\linewidth}
        \centering
        \caption{\textbf{Hi4D test.}}
        \label{tab:hi4d}
        \resizebox{\linewidth}{!}{%
        \begin{tabular}{l|cc}
        \toprule
        \multicolumn{1}{c|}{\multirow{2}{*}{\textbf{Data}}} & \multicolumn{1}{c}{\multirow{2}{*}{\small MPJPE ($\downarrow$ mm)}} & \multicolumn{1}{c}{\small Joint-PA} \\
         & & {\small MPJPE ($\downarrow$ mm)} \\ \midrule
        \rowcolor[HTML]{F2F2F2} Anny-One~\cite{bregier2025condimen} & 91.5 & 86.7 \\
        Anny-One + \cite{patel2025camerahmr} fits & 80.9 & 81.8 \\
        Anny-One + \textbf{our fits} & \textbf{80.1} & \textbf{79.5} \\
        $\Delta$ & {\color[HTML]{00B050} -11.4} & {\color[HTML]{00B050} -7.2} \\
        \bottomrule
        \end{tabular}
        }
    \end{subtable}
    \vspace{-2 em}
\end{table*}

%% file: sec/5_conclusion.tex
\section{Limitations}
\label{sec:limitations}
As an optimization-based method, our approach is highly dependent on the quality of both the initial mesh parameters and expert predictions. Fig.~\ref{fig:qual_neg} shows failure cases.
For example, errors in the experts --such as low-confidence keypoints or misclassified shape attributes--can prevent pose and shape convergence. Similarly, poor global-position initialization may cause optimization to stall or produce interpenetration when cues are weak. Future work could tackle this challenge, as explored previously by~\cite{muller2021self,mueller2023buddi,mihajlovic2025volumetricsmpl}. These issues highlight remaining challenges in complex multi-person scenes.

\section{Conclusion}
\label{sec:conclusion}
We introduced \ours, a robust multi-person optimization framework that jointly fits all individuals in the camera coordinate system using complementary expert signals, including VLM-derived semantic attributes. \ours improves spatial consistency, pose accuracy, and shape estimation across challenging benchmarks in all-age scenarios where standard per-person and traditional adult-only methods fail. We further demonstrated that \ours can generate high-quality pseudo-ground-truth annotations at scale, enabling the training of feedforward HMR models that achieve strong performance while predicting semantically meaningful shape parameters. Together, these results highlight the potential of expert-guided optimization to bridge the gap between adult-only models and real-world, all-age human reconstruction.

%% file: supplementary/supp.tex
\clearpage
\setcounter{page}{1}
\maketitlesupplementary

\begin{figure*}[!h]
    \centering
    \includegraphics[width=1.0\linewidth]{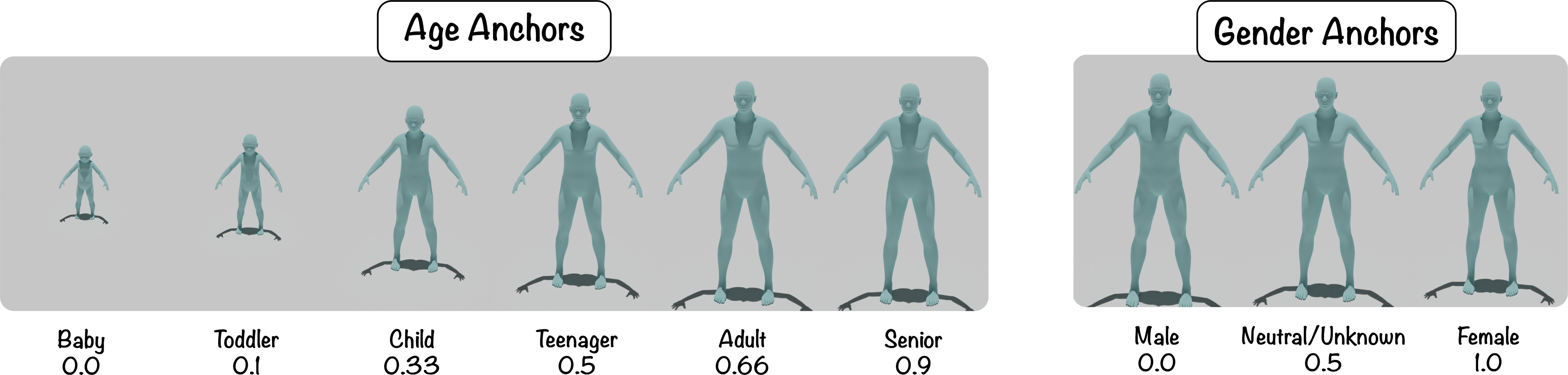}
 \caption{\textbf{Shape parameter anchors.} Examples of the text descriptors and Anny shape parameter values for each anchor. Left: Age mapping. Right: Gender mapping. Other shape parameters are set to 0.5 here for visualization. (Age set to 0.66 for gender).}
    \label{fig:anchors}
\end{figure*}

\section{Implementation details}
\label{supp:implementation}

\subsection{Anny model mapping}
\label{supp:anny_mapping}
We utilize the semantic shape space of the Anny model~\cite{bregier2025human} to propose a direct mapping from shape attribute descriptors to normalized shape values. This mapping inherently accounts for the body model interpolation described in the same work. Figure~\ref{fig:anchors} illustrates our complete mapping scheme for all experiments. To increase discriminative power, we supplement the age categories proposed in \cite{bev} by adding more resolution in the earlier ages where body shape changes rapidly. For gender, we set three equidistant anchors to better capture the continuous spectrum of visual genders.

\subsection{Pre-processing feedforward initializations}

\textbf{Multi-HMR.} As a bottom-up feed-forward model, Multi-HMR often struggles detecting certain people in the image. We particularly note how the non-maximum suppression process can remove highly occluded predictions. To recover these lost instances, we use the predicted nose keypoints from ViTPose to generate a mesh estimate for them.
We perform a mutual best assignment matching between nose keypoints and head detections. If a nose keypoint lacks a corresponding head prediction, we select the closest patch available. Overall, this strategy improves the recall on the test set from 74.0\% to 83.0\%. We note that when the initial head prediction is missing, the mesh estimate can be highly inaccurate, forcing the final optimization to rely heavily on other cues.

\textbf{From SMPL to Anny fits.}
Given a SMPL(X) mesh, we first convert it to the Anny topology using a vertex regressor~\cite{bregier2025human}.
We then estimate the corresponding Anny pose and shape parameters using a fast fitting algorithm similar to NLF~\cite{sarandi2024neural}.
The method alternates for 5 iterations between two steps: we independently fit the global orientation of each body part via a weighted Kabsch algorithm using Anny’s blend skinning weights, and then solve for the shape parameters with linear least squares.
Finally, we optionally refine the part rotations in a single forward pass along the kinematic tree.
The procedure is highly efficient, supports batch GPU execution, and processes 128 meshes in about 100 ms.

\subsection{Vision-Language Model prompting}
\label{supp:vlm_prompts}
\textbf{Prompts.} For estimating the shape attributes with a VLM we use the names of the anchors described in \ref{supp:anny_mapping}, below we show the prompts:

\input{supplementary/tables/prompts}

\input{supplementary/tables/ablation_prompting}

\textbf{Ablation on prompting strategies.}
Given our multi-person image setting, we studied the effect of varying the prompting strategy to the VLM when specifying individual subjects. Table~\ref{tab:vlm_prompting} shows the performance of different methods for utilizing the person's bounding box. The least effective strategy is the grounded approach, which provides the bounding box coordinates directly in the text prompt. Performance improves when we visually overlay the box in red directly onto the image. The most effective approaches involve spatially focusing the VLM, specifically by cropping the person's full body or cropping only their head. The head crop proved particularly useful for images featuring crowds or significant occlusions. We hypothesize that combining both the full body crop and the head crop would likely overcome the limitations of each approach, but at an increase in computational cost.

\subsection{Optimization parameters}
For our experiments we set all parameter learning rates to 0.01 except for the full body rotation $\theta$ and shape ($\beta$) which are set to 0.001. Across all stages $\lambda_{shape} = 10.0, \lambda_{2D} = 0.01$ and $\lambda_{dense} = 0.001$ with $\sigma = 100$. The first stage where we update only $\tau$ runs for 50 iterations for Multi-HMR and 200 iterations for CameraHMR with $\lambda_{init} = 10.0, \lambda_{depth} = 10.0$. The second stage, that updates $\{\tau, \phi, \beta \}$, runs for 100 iterations for both initializations, with $\lambda_{depth} = 50.0$, and we split $\lambda_{init}$ into the more granular $\lambda_{init_{\beta}} = \lambda_{init_{verts}} = 0.01, \lambda_{init_{\phi}} = 10.0$. The final stage (where all parameters are optimized) runs for 200 iterations with $\lambda_{depth} = 0.0$, $\lambda_{init_{\phi}} = \lambda_{init_{verts}} = 0.01, \lambda_{init_{\beta}} = 5.0$.

\subsection{}

\section{Experiment details}

\subsection{Metrics}
\textbf{Percentage of Correct Keypoints (PCK)}. To robustly account for missing keypoint detections, prior work often assigns a fixed “punishment value” to unmatched predictions when computing the MPJPE. However, such heuristics distort the numerical scale of the evaluation and can introduce undesirable incentives—e.g., a missed detection may be penalized less than an inaccurate prediction. For this reason, we complement MPJPE with the PCK metric, which measures the proportion of predicted keypoints falling within a predefined distance threshold from the ground truth. This formulation naturally handles both poor localization and missing detections without relying on arbitrary penalty terms, yielding a more interpretable and principled performance measure. Note that we follow the same evaluation procedure on Relative Human, but extend prior implementations to account for non-detected individuals.

\subsection{Relative Human}
\label{supp:rh_details}

\begin{figure}[!h]
    \centering
    \includegraphics[width=0.7\linewidth]{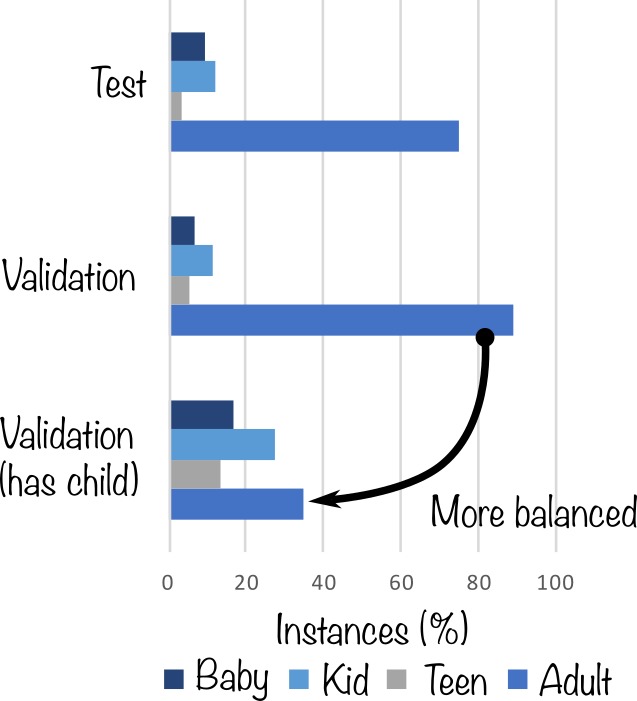}
    \caption{Age distribution across different subsets of the Relative Human dataset. We use the 'has child' subset for our ablation experiments.}
    \label{fig:rh_stats}
\end{figure}

\textbf{Ablation 'has child' subset.}
For the ablation experiments we report results on the validation 'has child' subset, composed by all the images that have at least one non-adult example. Figure~\ref{fig:rh_stats} show the resulting class distribution, where the 'has child' validation subset balances corrects the over-representation of adults. 

\textbf{Impact of detection.}
As an additional experiment we run \ours on the test set using ground-truth bounding boxes. Table~\ref{tab:bbox_exp} shows how the improvements from \ours are consistent when having ground-truth bounding boxes. We find that the ground-truth bounding boxes are helpful to recover occluded, back-facing or far away people in the image.

\input{supplementary/tables/bbox_exp}
\input{supplementary/tables/phmr}

\noindent \textbf{Additional results.} \ours also improves on the scene-aware method PromptHMR~\cite{wang2025prompthmr}. We use as prompts bounding boxes, segmentation mask, and find no change when including text descriptions, likely because age modeling is not part of the text training corpus of PromptHMR.

\begin{figure*}
    \centering
    \includegraphics[width=0.75\linewidth]{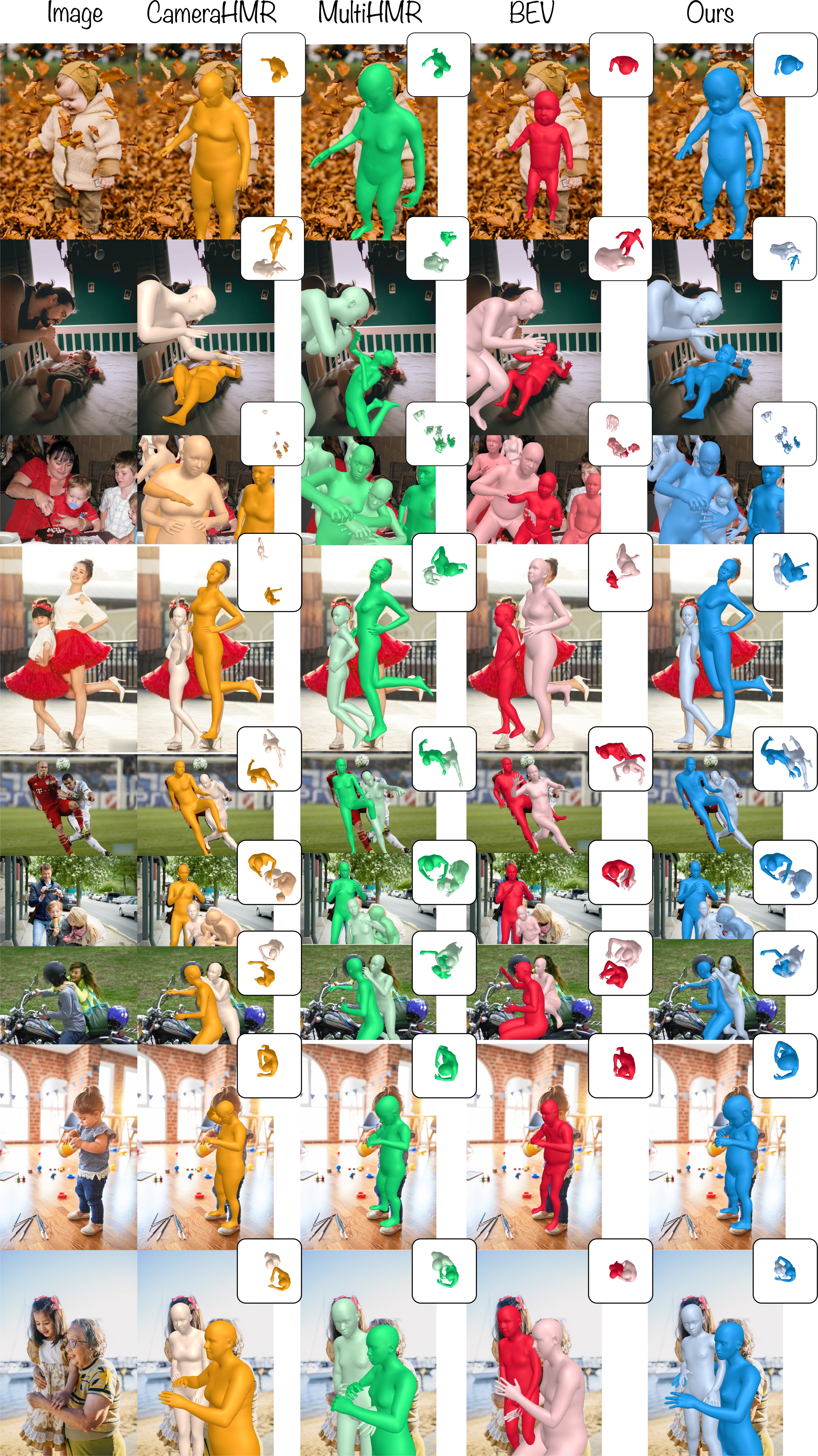}
    \caption{Additional examples of reconstructions with \ours (Ours) compared to SOTA on the Relative Human dataset.}
    \label{fig:supp_qual_pos}
\end{figure*}

\textbf{Additional qualitative results.} Figure~\ref{fig:supp_qual_pos} shows additional qualitative results on the Relative Human dataset.

\subsection{CMU Toddler}
\label{supp:cmu_details}
To the best of our knowledge, this is the only dataset from real captures with 3D ground-truth annotations for non-adults. We select the 5 sequences from the CMU Panoptic dataset~\cite{Joo_2017_TPAMI} that include a non-adult, specifically the sequences follow a toddler and 1 to 3 adults interacting, sometimes with objects. Following standard HMR evaluations on the 4 adult sequences~\cite{zanfir2018monocular,bev}, we evaluate all frames at 1 FPS, on 2 selected cameras and on the valid 3D joints (those in frame). For most sequences, the toddler is out of frame for a large part of the videos because of their height. As such, we select the following cameras to ensure the subject is in frame: Ian 1: $\{16, 21\}$, Ian 2 and Ian 5: $\{11, 21\}$, Ian 3: $\{11, 23\}$, Toddler 5: $\{15, 23\}$.

\begin{figure*}[!h]
    \centering
    \includegraphics[width=1.0\linewidth]{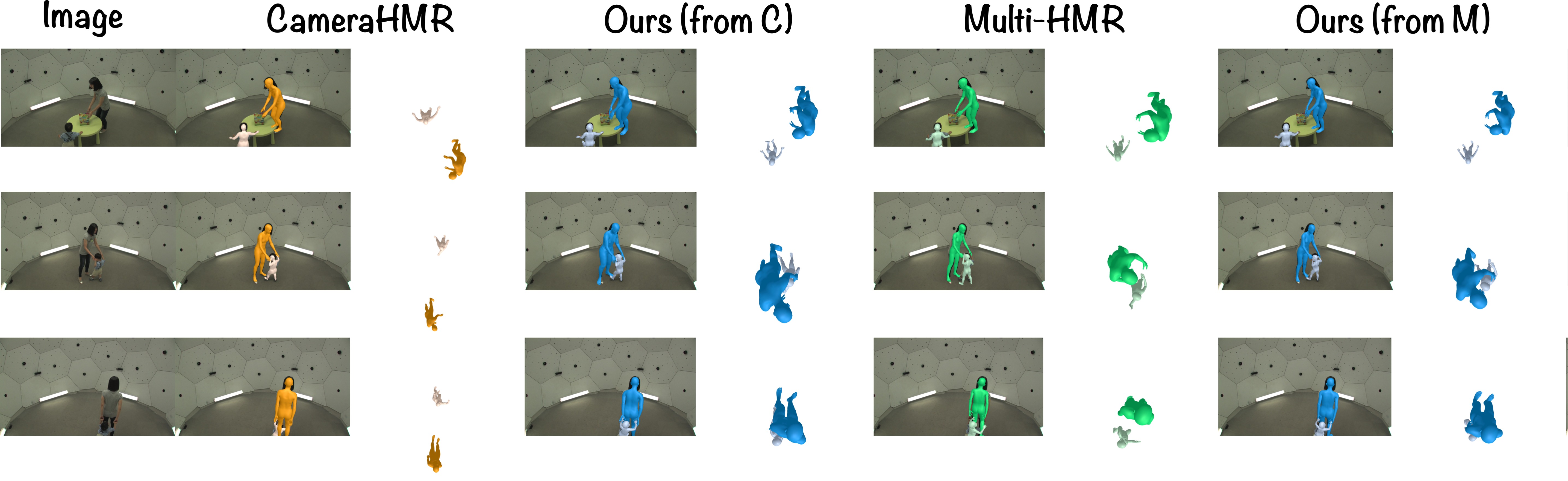}
    \caption{\textbf{Results on CMU toddler.} For each method we show the camera and top views.}
    \label{fig:cmu_pos}
\end{figure*}

Figure~\ref{fig:cmu_pos} shows examples of \ours applied to CMU Panoptic toddler sequences, using both initializations. Our observations are consistent with Table~\ref{tab:cmu}: Multi-HMR shows strong baseline performance, where our method refines relative positioning and pose. Conversely, for CameraHMR, the primary gain lies in relative positioning and shape.

\subsection{Training with pseudo-ground truth fits}
\begin{figure*}[!h]
    \centering
    \includegraphics[width=0.8\linewidth]{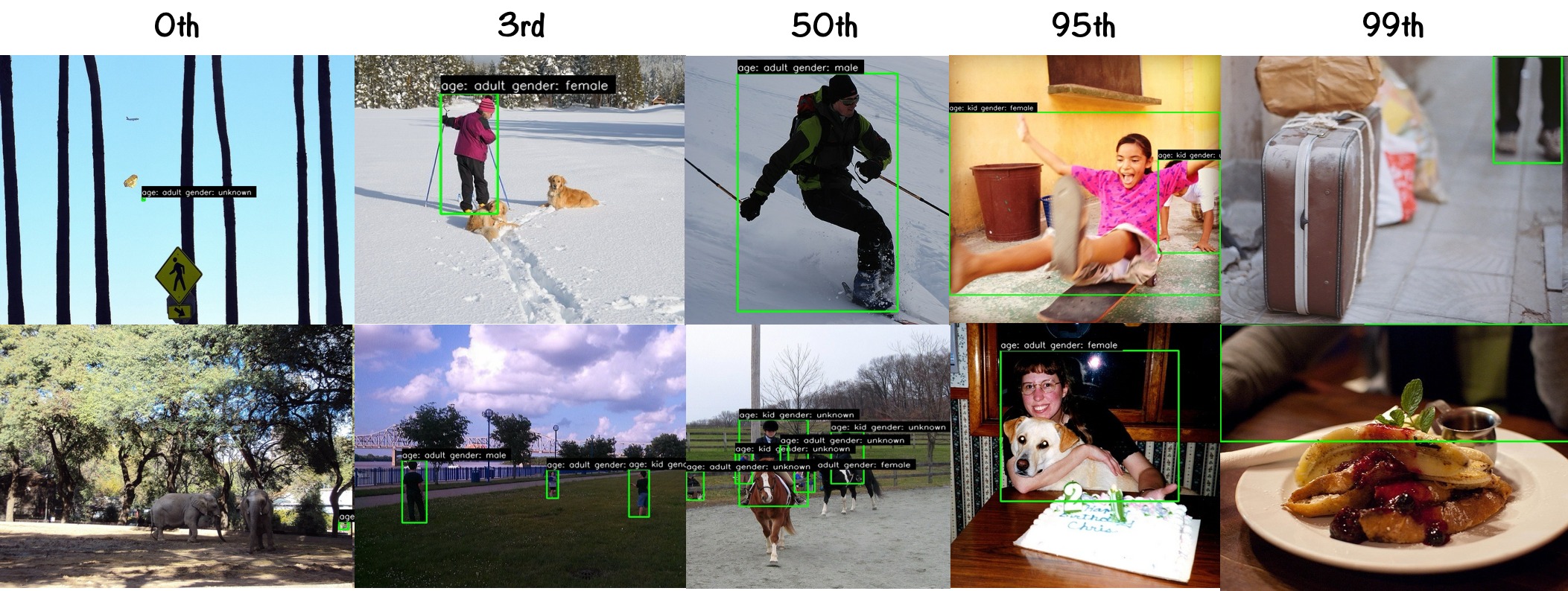}
    \caption{\textbf{Visualization of 2D Keypoint reprojection error percentiles.} We filter out fits that are not within the 3rd and 95th percentiles. We overlay the ground-truth bounding boxes and estimated age and gender.}
    \label{fig:error_percentiles}
\end{figure*}

\textbf{Fitting on MS-COCO.} We create pseudo- ground-truth fits for 30537 images in the MS-COCO train dataset using the ground-truth bounding boxes. Before training, we filter images with unsuccessful fits using the proxy of the 2D reprojection loss for all people. We select the images within the 3rd and 95th error percentiles, leaving 28039 after filtering. Figure~\ref{fig:error_percentiles} shows examples of the error percentiles. The lowest percentiles are likely due to not enough estimated keypoints, while the highest percentiles point to highly cropped people.

\begin{figure}[!h]
    \centering
    \includegraphics[width=1.0\linewidth]{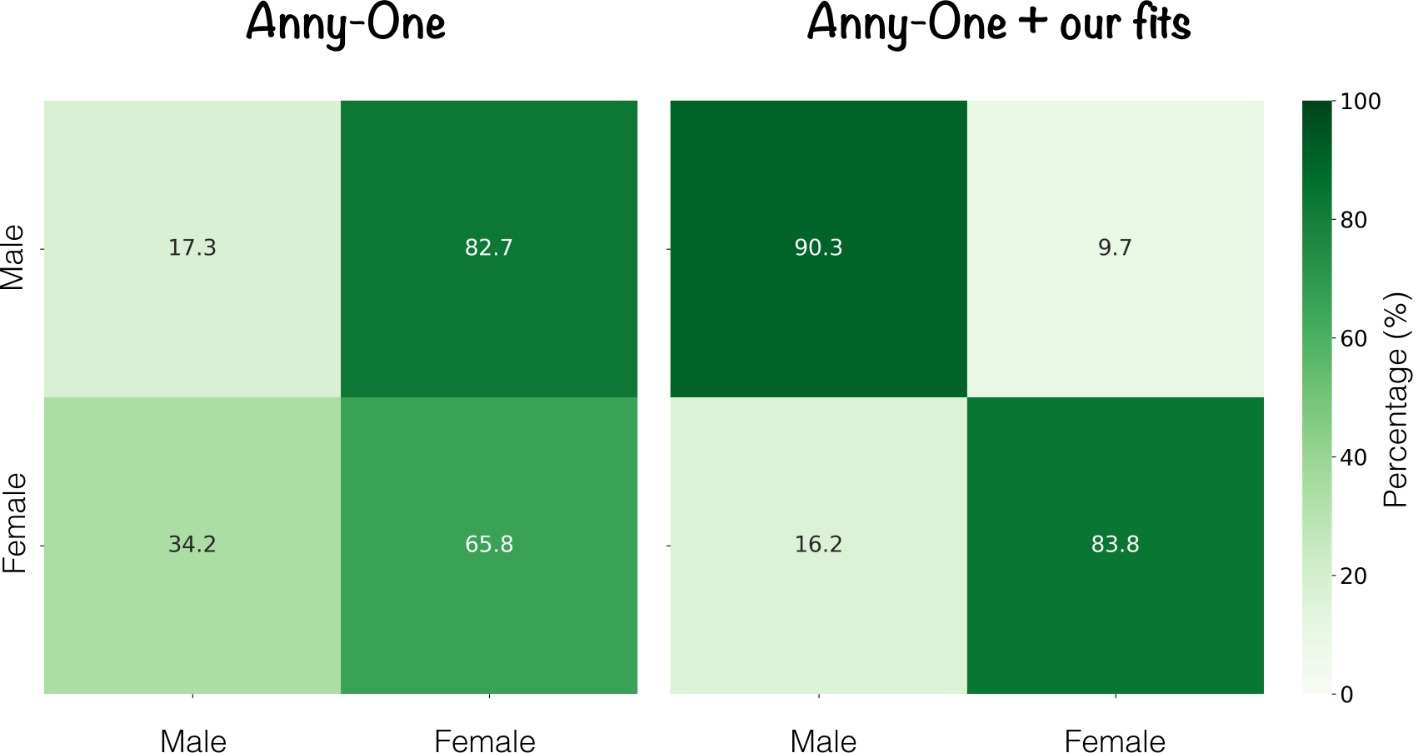}
    \caption{\textbf{Gender confusion matrix on Relative Human test.} Note how after retraining with our \ours fits, the model can accurately predict gender in an unseen dataset.}
    \label{fig:gender_conf}
\end{figure}

\textbf{Gender prediction.} Figure~\ref{fig:gender_conf} shows the effect of retraining with our fits. Similarly to Figure~\ref{fig:conf_mat}, the improved performance exemplifies how \ours can effectively distill information onto a retrained model.

\section{Beyond all-age shape estimation}
\begin{figure}[th]
    \centering
    \includegraphics[width=0.8\linewidth]{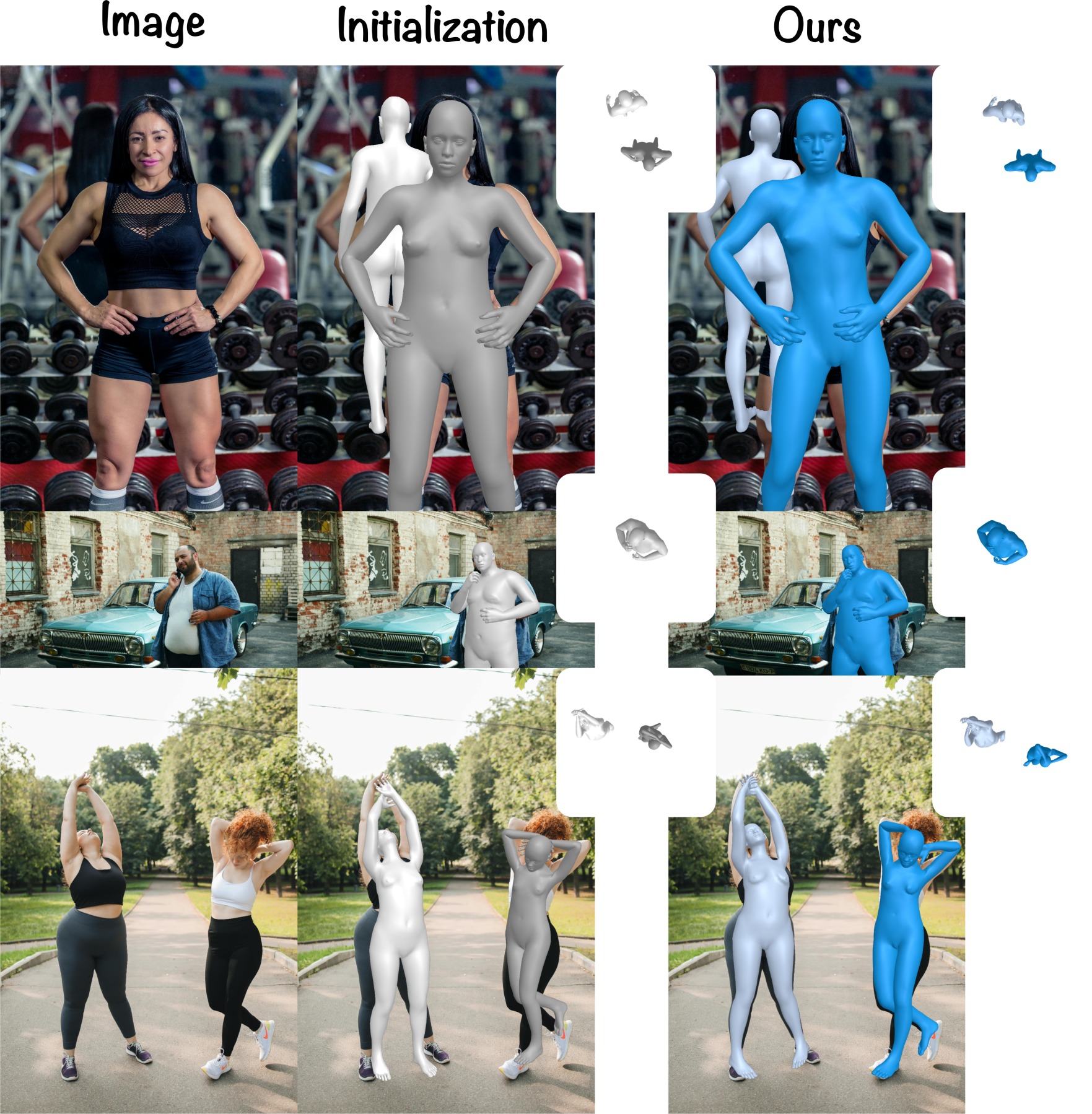}
    \caption{Optimization results from \ours demonstrating estimation of diverse weight and muscle shape attributes. Top: 'muscular' Mid: 'overweight'; and Bottom: 'overweight' and 'slim'.}
    \label{fig:body_shape}
\end{figure}

While not our primary objective, our shape estimation formulation generalizes to account for attributes beyond age and gender. As a case study, we explore diverse body shapes. To manage a diverse range of shapes with a compact categorization for VLM querying, we define a discrete mapping to weight and muscle attributes using the following anchors: $slim = \{muscle: 0.3, weight:0.3\}$, $average = \{muscle: 0.4, weight: 0.7\} $, $overweight = \{muscle: 0.1, weight: 0.8\}$, and $muscular=\{muscle: 0.9, weight: 0.7\}$. Figure~\ref{fig:body_shape} presents preliminary examples of how mapping the muscle and weight dimensions of the Anny body model improves shape estimation on in-the-wild images (sourced from Pexels~\cite{pexels}). These results suggest that \ours can be extended to align 3D reconstructions with other shape estimation settings.

\section{Limitations}
Our method integrates multiple expert predictions to guide the optimization for all-age human reconstruction, which enhances overall robustness. While this multi-expert strategy improves robustness, it also makes performance dependent on the accuracy of each expert. Errors in keypoints or depth under occlusion or extreme viewpoints can propagate through the optimization. Furthermore, age and gender estimates rely heavily on facial cues, making them unreliable for back-facing, occluded, or low-resolution subjects; inaccurate estimates can consequently bias the inferred shape. While the scarcity of child datasets (due to privacy and ethical constraints) limits comprehensive evaluation for younger age groups. Finally, while our method jointly optimizes all visible individuals in multi-person scenes, it does not explicitly handle interpenetration. Accounting for this could provide a valuable physical prior for the optimization and help prevent implausible overlaps between people.

%% file: supplementary/tables/prompts.tex
\begin{tcolorbox}[
    enhanced,
    colback=gray!10,
    colframe=SeabornBlue,
    boxrule=0.5pt,
    title=Gender estimation prompt,
    fonttitle=\sffamily,
    coltitle=white,
    arc=3pt,
    boxsep=5pt
]
\label{prompt:gender}
\ttfamily
"Look at the cropped image of a person and estimate their gender. Choose exactly one of the following categories: male, female, unknown.
Respond with only the category name.
For example, if the person is female, respond with 'female'."
\end{tcolorbox}

\begin{tcolorbox}[
    enhanced,
    colback=gray!10,
    colframe=SeabornBlue,
    boxrule=0.5pt,
    title=Age estimation prompt,
    fonttitle=\sffamily,
    coltitle=white,
    arc=3pt,
    boxsep=5pt
]
\ttfamily
"Look at the cropped image of a person and estimate their age group.
Choose one of the following categories according to the age in years: baby (0 to 1), toddler (2 to 3), kid (4 to 8), teen (8 to 16), adult (16+).
Respond with only the category name.
For example, if the person is a toddler, respond with 'toddler'."
\end{tcolorbox}

%% file: supplementary/tables/ablation_prompting.tex
\begin{table}[!th]
\centering
\caption{\textbf{Prompting strategies.} Results of varying prompting strategies for the VLM model (Qwen 2.5 VL) on the Relative Human validation 'has child' subset.
\colorbox[HTML]{FFF2CC}{\phantom{x}} indicates the default setting.
}
\label{tab:vlm_prompting}
\resizebox{\linewidth}{!}{%
\begin{tabular}{l|ccccc|c} 
\toprule
\multicolumn{1}{c |}{\multirow{2}{*}{\textbf{Method}}} & \multicolumn{5}{c |}{\textbf{Age} \small F1} & \textbf{Gender} \\
\multicolumn{1}{c |}{} & overall & adult & teen & kid & baby & F1 \\
\midrule
grounded & 40.42 & 21.96 & 45.83 & 52.53 & 41.35 & 70.38 \\
overlay & 57.41 & 82.56 & 38.46 & 64.47 & 44.16 & 89.85 \\
person crop & 66.69 & 85.77 & 50.63 & 70.25 & 60.12 & 91.22 \\
\rowcolor[HTML]{FFF2CC} head crop & 67.15 & 85.98 & 51.57 & 70.43 & 60.61 & 91.22 \\

\bottomrule
\end{tabular}
}
\end{table}

%% file: supplementary/tables/bbox_exp.tex
\begin{table}[th]
\centering
\caption{\textbf{Reconstruction with ground-truth bounding boxes.} $\Delta$: improvement over initialization. $\dagger$: ground-truth boxes used for inference.}
\label{tab:bbox_exp}
\resizebox{\linewidth}{!}{%
    \begin{tabular}{r|cccccccc} 
    \toprule
    \multicolumn{1}{r|}{\multirow{2}{*}{Method}} & \textbf{2D} ($\uparrow$) & \multicolumn{5}{c}{$\mathbf{PCRD^{0.2}}$ ($\uparrow$)} & \multicolumn{1}{c}{\textbf{Age} ($\uparrow$)} & \multicolumn{1}{c}{\textbf{Gender} ($\uparrow$)} \\
    \multicolumn{1}{c|}{} & $mPCKh^{0.6}$ & overall & adult & teen & kid & baby & F1 & F1 \\ \midrule

    \rowcolor[HTML]{F2F2F2} Multi-HMR~\cite{bregier2025human}$\dagger$ & 65.43 & 60.74 & 61.10 & 68.36 & 57.67 & 44.99 & 24.01 & 35.40 \\
    
    + \textbf{Ours} & 79.37 & 67.78 & 68.20 & 71.80 & 67.04 & 57.55 & 49.00 & 84.45 \\

    $\Delta$ & {\color[HTML]{00B050} +13.94}  & {\color[HTML]{00B050} +7.04} & {\color[HTML]{00B050} +7.1} & {\color[HTML]{00B050} +3.44} & {\color[HTML]{00B050} +9.37} & {\color[HTML]{00B050} +12.56} & {\color[HTML]{00B050} +24.99} & {\color[HTML]{00B050} +45.05} \\
    
    \rowcolor[HTML]{F2F2F2} CameraHMR~\cite{patel2025camerahmr}$\dagger$ & 64.47 & 59.7 & 59.86 & 67.49 & 47.21 & 31.16 & 0.00 & 0.00 \\
    
    + \textbf{Ours} & 82.02 & 66.32 & 66.45 & 72.96 & 61.94 & 53.20 & 44.08 & 83.34 \\

    $\Delta$ & {\color[HTML]{00B050} +17.55} & {\color[HTML]{00B050} +6.62} & {\color[HTML]{00B050} +6.59} & {\color[HTML]{00B050} +5.47} & {\color[HTML]{00B050} +14.73} & {\color[HTML]{00B050} +22.04} & {\color[HTML]{00B050} +44.08} & {\color[HTML]{00B050} +83.34} \\
    \bottomrule 
    \end{tabular}
    }
\end{table}

%% file: supplementary/tables/phmr.tex
\begin{table}{}
        \vspace{-0.5em}
        \centering
        \caption{\textbf{Additional \ours reconstruction results on Relative Human test}.}
        \vspace{-0.5em}
        \label{tab:results_phmr}
        \resizebox{\linewidth}{!}{%
        \begin{tabular}{r|cccccccc} 
        \toprule
        \multicolumn{1}{r|}{\multirow{2}{*}{Method}} & \textbf{2D} ($\uparrow$) & \multicolumn{5}{c}{$\mathbf{PCRD^{0.2}}$ ($\uparrow$)} & \multicolumn{1}{c}{\textbf{Age} ($\uparrow$)} & \multicolumn{1}{c}{\textbf{Gender} ($\uparrow$)} \\
        \multicolumn{1}{c|}{} & $mPCKh^{0.6}$ & overall & adult & teen & kid & baby & F1 & F1 \\ \midrule
        \rowcolor[HTML]{F2F2F2} PromptHMR~[56]
        & 67.70 & 60.15 & 60.80 & 61.90 & 47.48 & 29.06 & 0 & 0 \\
        + \textbf{Ours} & 81.84 & 67.88 & 68.61 & 70.08 & 66.56 & 52.88 & 48.73 & 81.72\\

        $\Delta$ & {\color[HTML]{00B050} +14.14}  & {\color[HTML]{00B050} +7.73} & {\color[HTML]{00B050} +7.81} & {\color[HTML]{00B050} + 8.18} & {\color[HTML]{00B050} +19.08} & {\color[HTML]{00B050} +23.82} & {\color[HTML]{00B050} +48.73} & {\color[HTML]{00B050} +81.72}\\
        \bottomrule
        \end{tabular}
        }%
        \vspace{-1.5em}
    \end{table}